\documentclass[runningheads]{llncs}

\usepackage[utf8]{inputenc} % allow utf-8 input
\usepackage[T1]{fontenc}    % use 8-bit T1 fonts
\usepackage{hyperref}       % hyperlinks
\usepackage{url}            % simple URL typesetting
\usepackage{booktabs}       % professional-quality tables
\usepackage{amsfonts}       % blackboard math symbols
\usepackage{nicefrac}       % compact symbols for 1/2, etc.
\usepackage{microtype}      % microtypography
\usepackage{xcolor}         % colors
\usepackage{amsmath}
\usepackage{graphicx}
\usepackage{rotating}
\usepackage{tabulary}
\usepackage{algorithm}
\usepackage{algorithmic}
\usepackage{subdef}
\usepackage{wrapfig,lipsum,booktabs}
\usepackage{subcaption}
\usepackage{todonotes}
\usepackage[misc]{ifsym}

\usepackage[numbers]{natbib}

\newcommand{\figref}[1]{Fig.~\ref{#1}}

\newcommand{\secref}[1]{Sec.~\ref{#1}}

\newcommand{\Appref}[1]{Appendix.~\ref{#1}}
\newcommand{\model}{\mbox{\textsc{Clinical}}}
\begin{document}

\title{CLINICAL: Targeted Active Learning for Imbalanced Medical Image Classification}
\titlerunning{CLINICAL: Targeted AL for Imbalanced Medical Image Classification}
\author{Suraj Kothawade\inst{1}$^{\textrm{\Letter}}$ \and
Atharv Savarkar\inst{2} \and
Venkat Iyer\inst{2} \and
Lakshman Tamil\inst{1} \and
Ganesh Ramakrishnan\inst{2}\and
Rishabh Iyer\inst{1}}
\authorrunning{S. Kothawade et al.}
\institute{University of Texas at Dallas, USA \and
% \email{suraj.kothawade@utdallas.edu}\\
Indian Institute of Technology, Bombay, India\\
\email{suraj.kothawade@utdallas.edu}}
\maketitle              % typeset the header of the contribution

\begin{abstract}
  Training deep learning models on medical datasets that perform well for all classes is a challenging task. It is often the case that a suboptimal performance is obtained on some classes due to the natural class imbalance issue that comes with medical data. An effective way to tackle this problem is by using \emph{targeted active learning}, where we iteratively add data points that belong to the rare classes, to the training data. However, existing active learning methods are ineffective in targeting rare classes in medical datasets. In this work, we propose \model\ (targeted a\textbf{C}tive \textbf{L}earning for \textbf{I}mbala\textbf{N}ced med\textbf{IC}al im\textbf{A}ge c\textbf{L}assification) a framework that uses submodular mutual information functions as acquisition functions to mine critical data points from rare classes. We apply our framework to a wide-array of medical imaging datasets on a variety of real-world class imbalance scenarios - namely, \emph{binary} imbalance and \emph{long-tail} imbalance. We show that \model\ outperforms the state-of-the-art active learning methods by acquiring a diverse set of data points that belong to the rare classes.
%   $\approx 10\% - 12\%$ on the rare classes accuracy and $\approx 4\% - 6\%$ on overall accuracy for Path-MNIST and Pneumonia-MNIST image classification datasets.
\end{abstract}

\section{Introduction}
%Medical datasets are imbalanced-para1. Give some imbalanced statis from the datasets we are using.
Owing to the advancement of deep learning, medical image classification has made tremendous advances in the past decade. However, medical datasets are naturally imbalanced at the class level, \emph{i.e.}, some classes are comparatively rarer than the others. For instance, cancerous classes are naturally rarer than non-cancerous ones. In such scenarios, the over-represented classes \emph{overpower} the training process and the model ends up learning a biased representation. Deploying such biased models results in incorrect predictions, which can be catastrophic and even lead to loss of life. Active learning (AL) is a promising solution to mitigate this imbalance in the training dataset. The goal of AL is to select data points from an unlabeled set for addition to the training dataset at an additional labeling cost. The model is then retrained with the new training set and the process is repeated. Reducing the labeling cost using the AL paradigm is crucial in domains like medical imaging, where labeling data requires expert supervision ({\em e.g.}, doctors),
%such as in the form of doctors, 
which makes the process extremely expensive. However, current AL methods are inefficient in selecting data points from the rare classes in medical image datasets. Broadly, they use acquisition functions that are either: i) based on the uncertainty scores of the model, which are used to select the top uncertain data points \cite{settles2009active}, or ii) based on diversity scores, where data points having diverse gradients are selected \cite{ash2020deep, sener2018active}. They mainly focus on improving the overall performance of the model, and thereby fail to target these rare yet critical classes. Unfortunately, this leads to a wastage of expensive labeling resources when the goal is to improve performance on these rare classes.

% \begin{figure}[h]
\begin{wrapfigure}{R}{0.50\textwidth}
\centering
\begin{subfigure}[b]{0.5\textwidth}
   \includegraphics[width=1\linewidth]{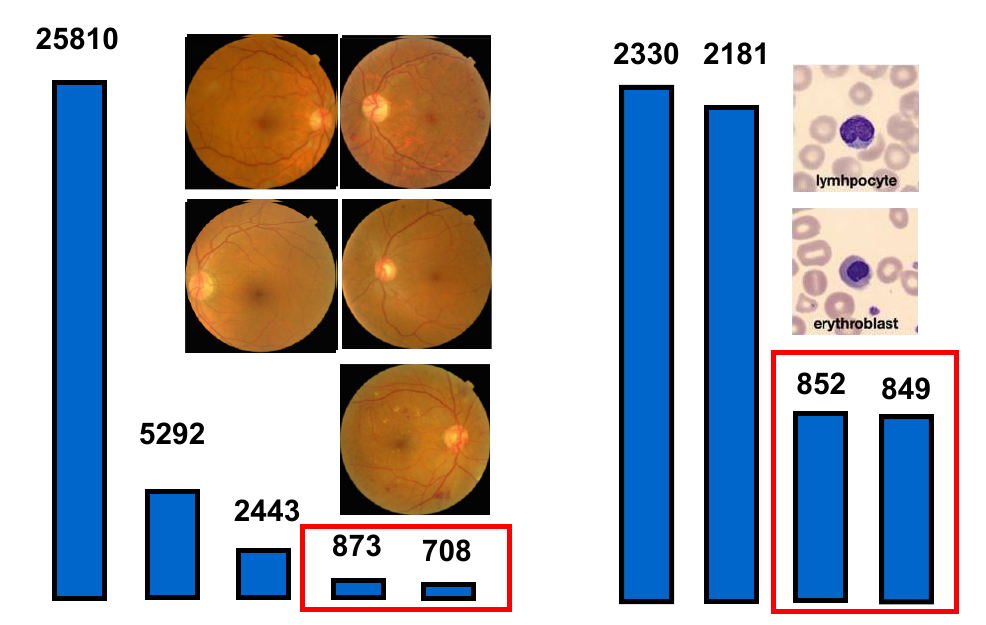}
%   \caption{}
   \label{fig:mot-longtail} 
\end{subfigure}
% \begin{subfigure}[b]{0.42\textwidth}
%   \includegraphics[width=1\linewidth]{plots/bloodmnist_binary.pdf}
%   \caption{}
%   \label{fig:mot-binary}
% \end{subfigure}
\caption{Motivating examples of two main class imbalance scenarios occurring in medical imaging. \textbf{Left:} Long-tail imbalance (Diabetic retinopathy grading from retinal images in APTOS-2019 \cite{aptos2019}). \textbf{Right:} Binary imbalance (Microscopic peripheral blood cell image classification in Blood-MNIST \cite{acevedo2020dataset}). Red boxes in both scenario denote targeted rare classes.}
\label{fig:motivating-scenarios}
% \vspace{-2ex}
% \end{figure}
\end{wrapfigure}
% \todo{The difference between long tail and binary imbalance is not clear from this figure. The first one could also be binary imbalance and the second one could also be long tail. What is the difference?}

%What are the solutions for this problem. AL is one solution and it can save costly labeling cost needed to hire doctors. Talk about AL methods and their disadvantages and why they cannot be used for rare classes scenario. Explain why our method is a better method
%We conisder two types of imbalance scenarios in our work - binary, longtail. Point to a figure, and cite some papers for both cases.
In this work, we consider two types of class imbalance that recur in a wide array of medical imaging datasets. The first scenario is \textit{binary} imbalance, where a subset of classes is rare/infrequent and the remaining subset is relatively frequent. The second scenario is that of \textit{long-tail} imbalance, where the frequency of data points from each class keeps \emph{steeply} reducing as we go from the most frequent class to the rarest class (see \figref{fig:motivating-scenarios}). Such class imbalance scenarios are particularly challenging in the medical imaging domain since there exist subtle differences which are barely visually evident (see \figref{fig:motivating-scenarios}). In \secref{sec:our_method}, we discuss \model, a targeted active learning algorithm that acquires a subset by maximizing the submodular mutual information with a set of \emph{misclassified} data points from the rare classes. This enables us to focus on data points from the unlabeled set that are critical and belong to the rare classes.

\subsection{Related work} \label{sec:realted_work}

%entropy, margin, least-conf
\textbf{Uncertainty based Active Learning.} Uncertainty based methods aim to select the most uncertain data points according to a model for labeling. The most common techniques are - 1) \textsc{Entropy} \cite{settles2009active} selects data points with maximum entropy, 2) \textsc{Least Confidence} \cite{wang2014new} selects data points with the lowest confidence, and 3) \textsc{Margin} \cite{roth2006margin} selects data points such that the difference between the top two predictions is minimum.
 
%badge, coreset, VAR, batchbald
\noindent \textbf{Diversity based Active Learning.} The main drawback of uncertainty based methods is that they lack diversity within the acquired subset. To mitigate this, a number of approaches have proposed to incorporate diversity. The \textsc{Coreset} method \cite{sener2018active} minimizes a coreset loss to form coresets that represent the geometric structure of the original dataset. They do so using a greedy \textit{k}-center clustering. A recent approach called \textsc{Badge} \cite{ash2020deep} uses the last linear layer gradients to represent data points and runs \textsc{K-means++} \cite{kmeansplus} to obtain centers that have a high gradient magnitude. The centers being representative and having high gradient magnitude ensures uncertainty and diversity at the same time. However, for batch AL, \textsc{Badge} models diversity and uncertainty only within the batch and \emph{not} across all batches. Another method, \textsc{BatchBald} \cite{kirsch2019batchbald} requires a large number of Monte Carlo dropout samples to obtain significant mutual information which limits its application to medical domains where data is scarce.
% \todo{Talk about and cite fass as well.}

%glister, gradmatch, similar, prism
\noindent \textbf{Class Imbalanced and Personalized Active Learning.}
%mention that Similar does it only as a POC for binary imbalance on non-real world datasets
Closely related to our method \model, are methods which optimize an objective that involves a held-out set. \textsc{GradMatch} \cite{killamsetty2021grad} uses an orthogonal matching pursuit algorithm to select a subset whose gradient closely matches the gradient of a validation set. Another method, \textsc{Glister-Active} \cite{killamsetty2020glister} formulates an acquisition function that maximizes the log-likelihood on a held-out validation set. We adopt \textsc{GradMatch} and \textsc{Glister-Active} as baselines that \emph{targets} rare classes in our class imbalance setting and refer to it \textsc{T-GradMatch} and \textsc{T-Glister} in \secref{sec:experiments}. Recently, \cite{kothawade2021similar} proposed the use of submodular information measures for active learning in realistic scenarios, while \cite{kothawade2021talisman} used them to find rare objects in an autonomous driving object detection dataset. Finally, \cite{kothyari2021personalizing} use the submodular mutual information functions (used here) for personalized speech recognition. Our proposed method uses the submodular mutual information to target selecting data points from the rare classes via using a small set of \emph{misclassified} data points as exemplars, which makes our method applicable to binary as well as long-tail imbalance scenarios.

\vspace{-2ex}

\subsection{Our contributions} \label{sec:contributions}
%1. we emphasize the issue of binary and longtail imbalance in medical datasets leading to poor performance on rare yet critical classes and wasted labeling efforts.
%2. we provide a novel AL framework that can be applied to any kind of imbalance
%3. we demonstrate the effectiveness of our framework on a diverse set of tasks and modalities, namely, 1)...2)... Furthermore, we show that Clinical outperforms the state-of-the-art AL methods by \approx X% -Y% in terms of average rare classes accuracy for binary imbalance scenarios and by \approx X% -Y% for long-tail imbalance scenarios.
%4. We provides insights on uncovering classes that are unknown unknowns using conditional gain functions.

We summarize our contributions as follows: \textbf{1)} We emphasize on the issue of binary and long-tail class imbalance in medical datasets that leads to poor performance on rare yet critical classes. \textbf{2)} Given the limitations of current AL methods on medical datasets, we propose \model, a novel AL framework that can be applied to any class imbalance scenario. \textbf{3)} We demonstrate the effectiveness of our framework for a diverse set of image classification tasks and modalities on Pneumonia-MNIST \cite{kermany2018identifying}, Path-MNIST \cite{kather2019predicting}, Blood-MNIST \cite{acevedo2020dataset}, APTOS-2019 \cite{aptos2019}, and ISIC-2018 \cite{codella2019skin} datasets. Furthermore, we show that \model\ outperforms the state-of-the-art AL methods by up to $\approx 6\% - 10\%$ on an average in terms of the average rare classes accuracy for binary imbalance scenarios and long-tail imbalance scenarios. \textbf{4)} We provide valuable insights about the \emph{choice} of submodular functions to be used for subset selection based on the \emph{modality} of medical data. 
% Lastly, \textbf{5)} We provide a strategy for uncovering classes that are unknown unknowns using conditional gain functions.
\vspace{-2ex}

\section{Preliminaries} \label{sec:preliminaries}

\textbf{Submodular Functions: } We let $\Vcal$ denote the \emph{ground-set} of $n$ data points $\Vcal = \{1, 2, 3,...,n \}$ and a set function $f:
 2^{\Vcal} \xrightarrow{} \mathbb{R}$.  
 The function $f$ is submodular~\citep{fujishige2005submodular}  if it satisfies diminishing returns, namely $f(j | \Xcal) \geq f(j | \Ycal)$ for all $\Xcal \subseteq \Ycal \subseteq \Vcal, j \notin \Ycal$.
Facility location, graph cut, log determinants, {\em etc.} are some examples~\citep{iyer2015submodular}. 
% Due to  close connections between submodularity and entropy, submodular functions can also be viewed as \emph{information functions}~\citep{zhang1998characterization}. Submodularity ensures that a greedy algorithm achieves bounded approximation factor when maximized~\citep{nemhauser1978analysis}.

\noindent \textbf{Submodular Mutual Information (\textsc{Smi}):} Given a set of items $\Acal, \Qcal \subseteq \Vcal$, the submodular mutual information (MI)~\citep{levin2020online,iyer2020submodular} is defined as $I_f(\Acal; \Qcal) = f(\Acal) + f(\Qcal) - f(\Acal \cup \Qcal)$. Intuitively, this measures the similarity between $\Qcal$ and $\Acal$ and we refer to $\Qcal$ as the query set. \cite{kothawade2021prism} extend \textsc{Smi} to handle the case when the \emph{target} can come from a different set $\Vcal'$ apart from the ground set $\Vcal$. In the context of imbalanced medical image classification, $\Vcal$ is the source set of images and the query set $\Qcal$ is the target set containing the rare class images.
% Let $\Omega  = \Vcal \cup \Vcal^{\prime}$. We define a set function $f: 2^{\Omega} \rightarrow \mathbb{R}$. Although $f$ is defined on $\Omega$, the discrete optimization problem will only be defined on subsets $\Acal \subseteq \Vcal$. 
To find an optimal subset given a query set $\Qcal \subseteq \Vcal^{\prime}$, we can define $g_{\Qcal}(\Acal) = I_f(\Acal; \Qcal)$, $\Acal \subseteq \Vcal$ and maximize the same.

\vspace{-2ex}
\subsection{Examples of \textsc{Smi} functions}
For targeted active learning, we use the recently introduced \textsc{Smi} functions in~\cite{iyer2020submodular, levin2020online} and their extensions introduced in \cite{kothawade2021prism} as acquisition functions. 
For any two data points $i \in \Vcal$ and $j \in \Qcal$, let $s_{ij}$ denote the similarity between them.

\noindent\textbf{Graph Cut MI (\textsc{Gcmi}):} The \textsc{Smi} instantiation of graph-cut (\textsc{Gcmi}) is defined as: $I_{GC}(\Acal;\Qcal)=2\sum_{i \in \Acal} \sum_{j \in \Qcal} s_{ij}$.
% \small{
% \begin{align} \label{eq:GCMI}
% I_{GC}(\Acal;\Qcal)=2\sum_{i \in \Acal} \sum_{j \in \Qcal} s_{ij}
% \end{align}}
% \normalsize
Since maximizing \textsc{Gcmi} maximizes the joint pairwise sum with the query set, it will lead to a summary similar to the query set $Q$. In fact, specific instantiations of \textsc{Gcmi} have been intuitively used for query-focused summarization for videos ~\cite{vasudevan2017query} and documents ~\cite{lin2012submodularity, li2012multi}. 

%For a set $D \subseteq (V \cup Q)$, the SMI instantiation of facility locations can take the following two forms: 

\noindent\textbf{Facility Location MI (\textsc{Flmi}):} We consider two variants of \textsc{Flmi}. The first variant is defined over $\Vcal$(\textsc{Flvmi}), the \textsc{Smi} instantiation can be defined as: $I_{FLV}(\Acal;\Qcal)=\sum_{i \in \Vcal}\min(\max_{j \in \Acal}s_{ij},  \max_{j \in \Qcal}s_{ij})$.
% \small{
% \begin{align} \label{eq:FL1MI}
% I_{FL1}(\Acal;\Qcal)=\sum_{i \in \Vcal}\min(\max_{j \in \Acal}s_{ij}, \eta \max_{j \in \Qcal}s_{ij})
% \end{align}}
% \normalsize
The first term in the min(.) of \textsc{Flvmi} models diversity, and the second term models query relevance. 
% An increase in the value of $\eta$ causes the resulting summary to become more relevant to the query. 

For the second variant, which is defined over $\Qcal$ (\textsc{Flqmi}), the \textsc{Smi} instantiation can be defined as: $I_{FLQ}(\Acal;\Qcal)=\sum_{i \in \Qcal} \max_{j \in \Acal} s_{ij} + \sum_{i \in \Acal} \max_{j \in \Qcal} s_{ij}$. \textsc{Flqmi} is very intuitive for query relevance as well. It measures the representation of data points that are the most relevant to the query set and vice versa. 
% It can also be thought of as a bidirectional representation score.

% \noindent\textbf{Facility Location MI - V2:} In the V2 variant, we set $D$ to be $V \cup Q$. The SMI instantiation of FL2MI can be defined as: $I_{FL2}(\Acal;\Qcal)=\sum_{i \in \Qcal} \max_{j \in \Acal} s_{ij} + \eta\sum_{i \in \Acal} \max_{j \in \Qcal} s_{ij}$.
% \small{
% \begin{align} \label{eq:FL2MI}
% I_{FL2}(\Acal;\Qcal)=\sum_{i \in \Qcal} \max_{j \in \Acal} s_{ij} + \eta\sum_{i \in \Acal} \max_{j \in \Qcal} s_{ij}
% \end{align}}
% \normalsize

\noindent\textbf{Log Determinant MI (\textsc{Logdetmi}):} The \textsc{Smi} instantiation of \textsc{Logdetmi} can be defined as: $I_{LogDet}(\Acal;\Qcal)=\log\det(S_{\Acal}) -\log\det(S_{\Acal} - S_{\Acal,\Qcal}S_{\Qcal}^{-1}S_{\Acal,\Qcal}^T)$.
% \small{
% \begin{align} \label{eq:logdetMI}
% I_{LogDet}(\Acal;\Qcal)=\log\det(S_{\Acal}) -\log\det(S_{\Acal} - \eta^2 S_{\Acal,\Qcal}S_{\Qcal}^{-1}S_{\Acal,\Qcal}^T)
% \end{align}}
% \normalsize
$S_{\Acal, \Qcal}$ denotes the cross-similarity matrix between the items in sets $\Acal$ and $\Qcal$. 

% The similarity matrix in constructed in such a way that the cross-similarity between $\Acal$ and $\Qcal$ is multiplied by $\eta$ to control the trade-off between query-relevance and diversity.

\section{\model: Our Targeted Active Learning framework for Binary and Long-tail Imbalance} \label{sec:our_method}

\begin{figure*}
% \hspace*{-2cm}
\includegraphics[width = \textwidth]{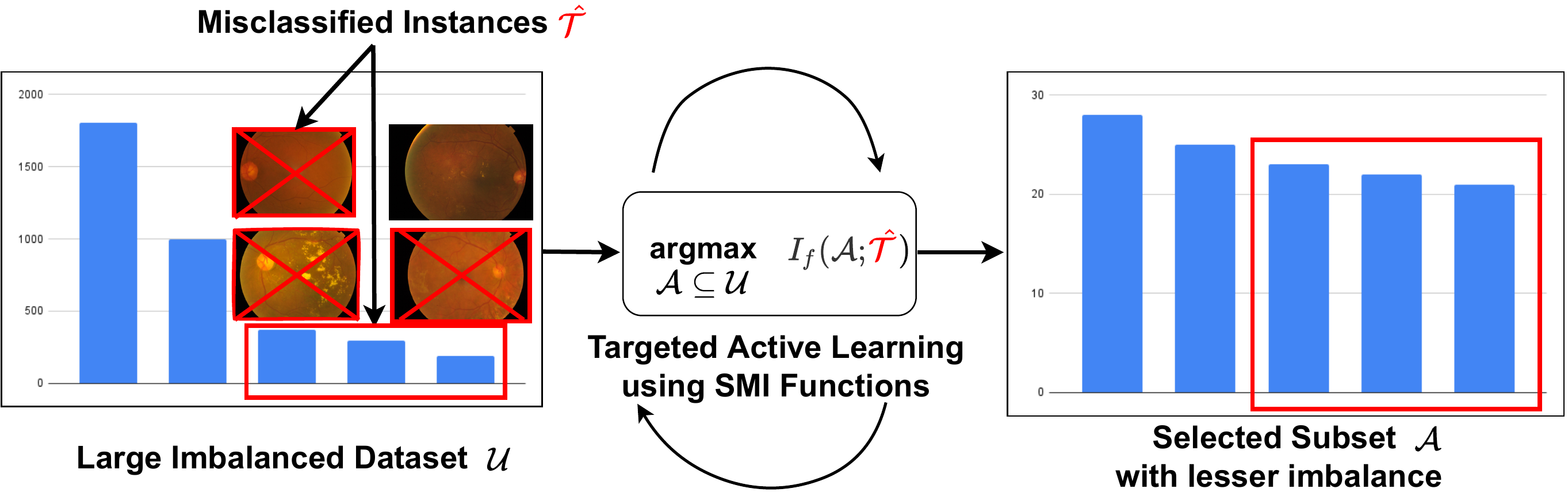}
\caption{The \model\ framework. We use a set of misclassfied instances $\hat{\Tcal}$ as the query set $\Qcal$ in the SMI function. We then maximize $I_f(\Acal; \hat{\Tcal})$ in an AL loop to target the imbalance and gradually mine data points from the rare classes.}
\vspace{-4ex}
\label{fig:clinical_arch}
\end{figure*}

% \subsection{Binary Imbalance}
In this section, we propose our targeted active learning framework, \model\  (see \figref{fig:clinical_arch}), and show how it can be applied to datasets with class imbalance. Concretely, we apply the \textsc{Smi} functions as acquisition functions for improving a model's accuracy on rare classes at a given additional labeling cost ($B$ instances) without compromising on the overall accuracy. The main idea in \model, is to use \emph{only the misclassified} data points from a held-out target set $\Tcal$ containing data points from the rare classes. Let $\hat{\Tcal} \subseteq \Tcal$ be the subset of misclassified data points. Then, we optimize the \textsc{Smi} function $I_f(\Acal; \hat{\Tcal})$ using a greedy strategy \cite{mirzasoleiman2015lazier}.  \looseness-1

% \begin{align}\label{eq:SMI-al}
% \max_{\Acal \subseteq \Ucal, |\Acal| \leq B} I_f(\Acal; \hat{\Tcal})    
% \end{align}

Note that since $\hat{\Tcal}$ contains only the misclassified data points, it would contain more data points from classes that are comparatively \emph{rarer} or the worst performing. Moreover, $\hat{\Tcal}$ is updated in every AL round, this mechanism helps the \textsc{Smi} functions to focus on classes that require the most attention. For instance, in the long-tail imbalance scenario (see \figref{fig:motivating-scenarios}), \model\ would focus more on the tail classes in the initial rounds of AL. Next, we discuss the \model\ algorithm in detail: \looseness-1

\vspace{0.2cm}

\noindent \textbf{Algorithm:} Let $\Lcal$ be an initial training set of labeled instances and $\Tcal$ be the target set containing examples from the rare classes. Let $\Ucal$ be a large unlabeled dataset and $\Mcal$ be the trained model using $\Lcal$. Next, we compute $\hat{\Tcal}$ as the subset of data points from $\Tcal$ that were misclassified by $\Mcal$. Using last layer gradients as a representation for each data point which are extracted from $\Mcal$, we compute similarity kernels of elements within $\Ucal$, within $\hat{\Tcal}$ and between $\Ucal$ and $\hat{\Tcal}$ to instantiate an \textsc{Smi} function $I_f(\Acal; \hat{\Tcal})$ and maximize it to compute an optimal subset $\Acal \subseteq \Ucal$ of size $B$ given $\hat{\Tcal}$ as target (query) set. We then augment $\Lcal$ with labeled $\Acal$ (i.e. $L(\Acal)$) and re-train the model to improve the model on the rare classes. We summarize \model\ in Algorithm~\ref{algo:clinical} and discuss its scalability aspects in \Appref{app:scalability}. \looseness-1

\vspace{-3ex}
\begin{algorithm}%[H]
\begin{algorithmic}[1]
\REQUIRE Initial Labeled set of data points: $\Lcal$, unlabeled dataset: $\Ucal$, target set: $\Tcal$, Loss function $\Hcal$ for learning model $\Mcal$, batch size: $B$, number of selection rounds: $N$ \\
\FOR{selection round $i = 1:N$}
\STATE Train $\Mcal$ with loss $\Hcal$ on the current labeled set $\Lcal$ and obtain parameters $\theta_i$
\STATE Compute $\hat{\Tcal} \subseteq \Tcal$ that were misclassified by the trained model $\Mcal$
\STATE Use $\Mcal_{\theta_i}$ to compute gradients using hypothesized labels $\{\nabla_{\theta} \mathcal H(x_j, \hat{y_j}, \theta), \forall j \in \Ucal\}$ and obtain a pairwise similarity matrix $X$. \textcolor{blue}{\{where $X_{ij} = \langle \nabla_{\theta} \Hcal_i(\theta), \nabla_{\theta} \Hcal_j(\theta) \rangle$\}}
\STATE Instantiate a submodular function $f$ based on $X$.
\STATE $\Acal_i \leftarrow \mbox{argmax}_{\Acal \subseteq \Ucal, |\Acal | \leq B}  I_f(\Acal; \hat{\Tcal})$
\STATE Get labels $L(\Acal_i)$ for batch $\Acal_i$, and $\Lcal \leftarrow \Lcal \cup L(\Acal_i)$, $\Ucal \leftarrow \Ucal - \Acal_i$
\STATE $\Tcal \leftarrow \Tcal \cup \Acal_i^\Tcal$, augment $\Tcal$ with new data points that belong to target classes. 
\ENDFOR
\STATE Return trained model $\Mcal$ and parameters $\theta_N$.
\end{algorithmic}
\caption{\model: Targeted AL for binary and long-tail imbalance}
\label{algo:clinical}
\end{algorithm}
\vspace{-7ex}
% \subsection{Long-tail Imbalance}

%Present the approximation to the min-max algorithm

\section{Experiments} \label{sec:experiments}
In this section, we evaluate the effectiveness of \model\ on binary imbalance (\secref{sec:exp_binary_imb}) and long-tail imbalance (\secref{sec:exp_longtail_imb}) scenarios. We do so by comparing the accuracy and class selections of various \textsc{Smi} functions with the existing state-of-the-art AL approaches. In our experiments, we observe that different \textsc{Smi} functions outperform exisiting approaches depending on the modality of the medical data. We show that the choice of the \textsc{Smi} based acqusition function is imperative and varies based on the imbalance scenario and the modality of medical data. Furthermore, in \Appref{app:pen_matrices}, we provide penalty matrices which show that \model\ statistically significantly outperforms the existing methods in all scenarios for multiple modalities.

\noindent \textbf{Baselines in all scenarios.} We compare the performance on \model\ against a variety of state-of-the-art uncertainty, diversity and targeted selection methods. The uncertainty based methods include \textsc{Entropy}, \textsc{Least Confidence} (\textsc{Least-Conf}), and \textsc{Margin}. The diversity based methods include \textsc{Coreset} and \textsc{Badge}. The targeted selection methods include \textsc{T-Glister} and \textsc{T-GradMatch}. We discuss the details of all baselines in \secref{sec:realted_work}. For a fair comparison with \model, we use the same target set of misclassified data points $\hat{\Tcal}$ as the held out validation set used in \textsc{T-Glister} and \textsc{T-GradMatch}. Lastly, we compare with random sampling (\textsc{Random}). \looseness-1

\noindent \textbf{Experimental setup:} We use the same training procedure and hyperparameters for all AL methods to ensure a fair comparison. For all experiments, we train a ResNet-18 \cite{he2016deep} model using an SGD optimizer with an initial learning rate of 0.001, the momentum of 0.9, and a weight decay of 5e-4. For each AL round, the weights are reinitialized using Xavier initialization and the model is trained till 99\% training accuracy. The learning rate is decayed using cosine annealing \cite{loshchilov2016sgdr} in every epoch. We run each experiment $5 \times$ on a V100 GPU and provide the error bars (std deviation). We discuss dataset splits for each our experiments below and provide more details in \Appref{app:dataset_details}.\\

\begin{figure*}[h]
\centering
\includegraphics[width = 12cm, height=1cm]{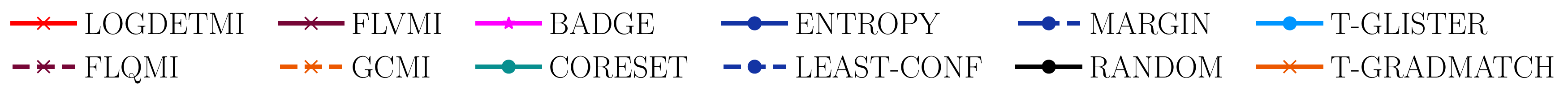}
% \centering
\begin{subfigure}[]{0.33\textwidth}
\includegraphics[width = \textwidth]{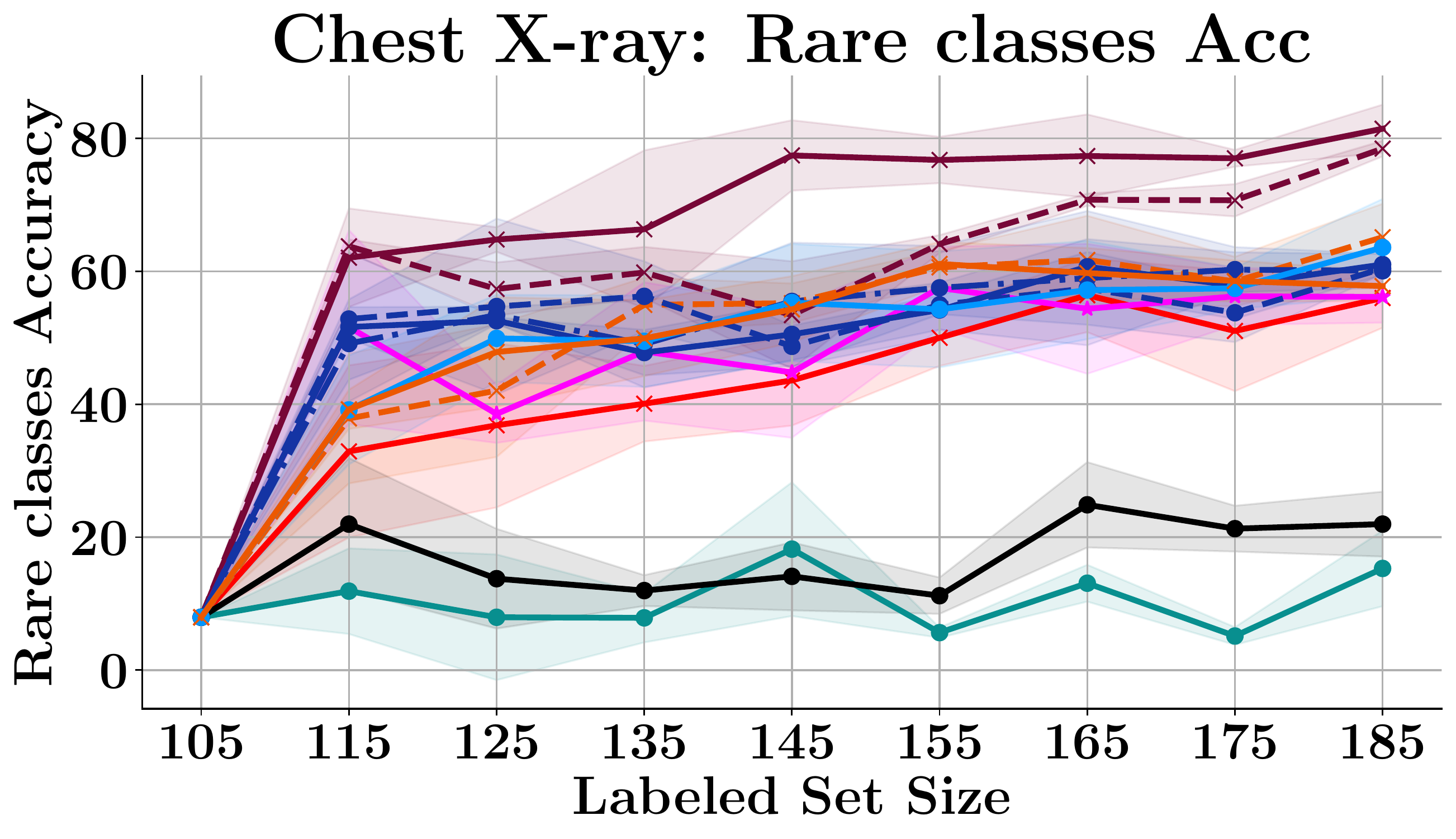}
% \caption{CMI vs Baselines (CIFAR-10)}
% \caption{}
\end{subfigure}
\begin{subfigure}[]{0.33\textwidth}
\includegraphics[width = \textwidth]{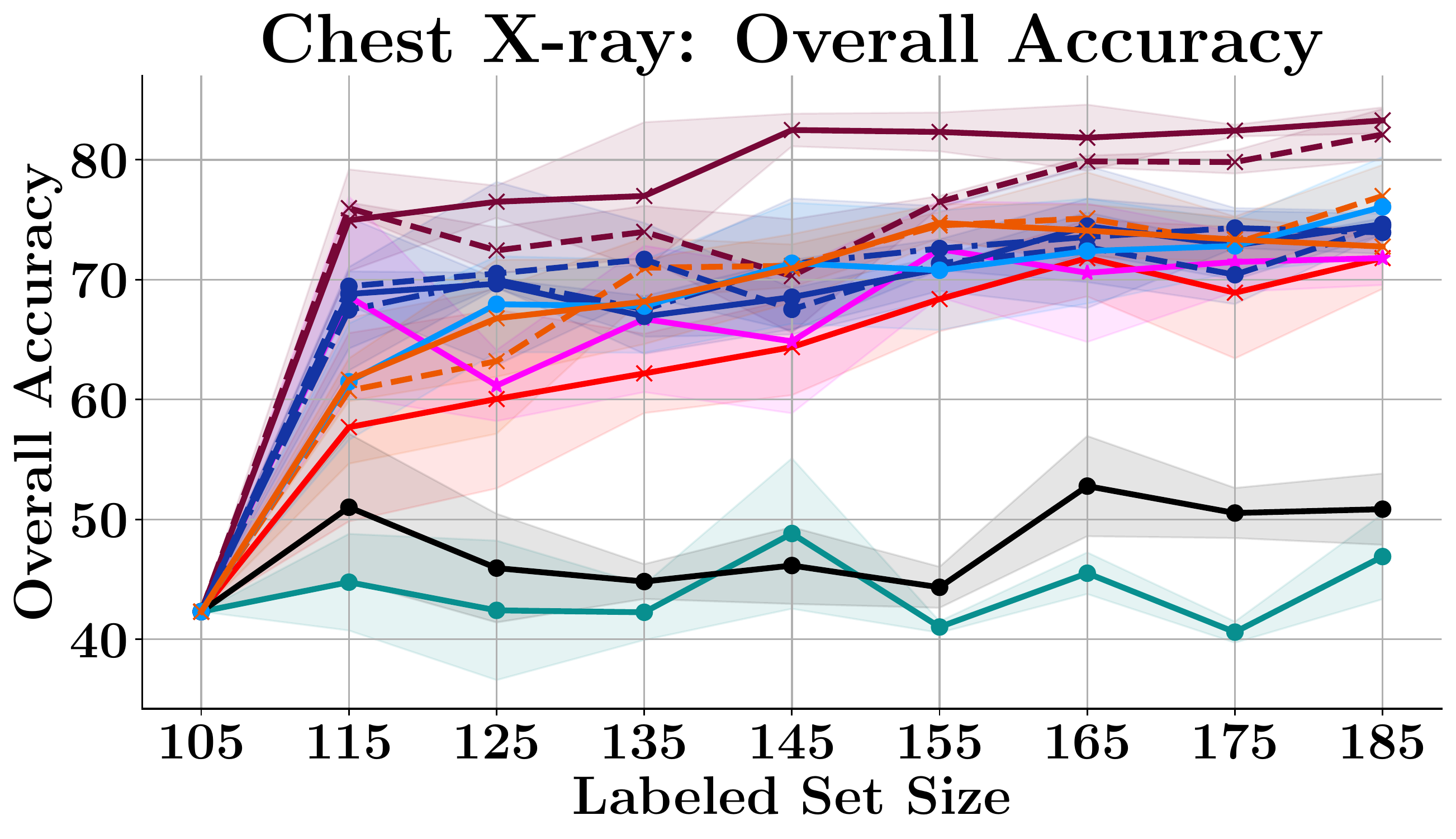}
% \caption{}
\end{subfigure}
\begin{subfigure}[]{0.32\textwidth}
\includegraphics[width = \textwidth]{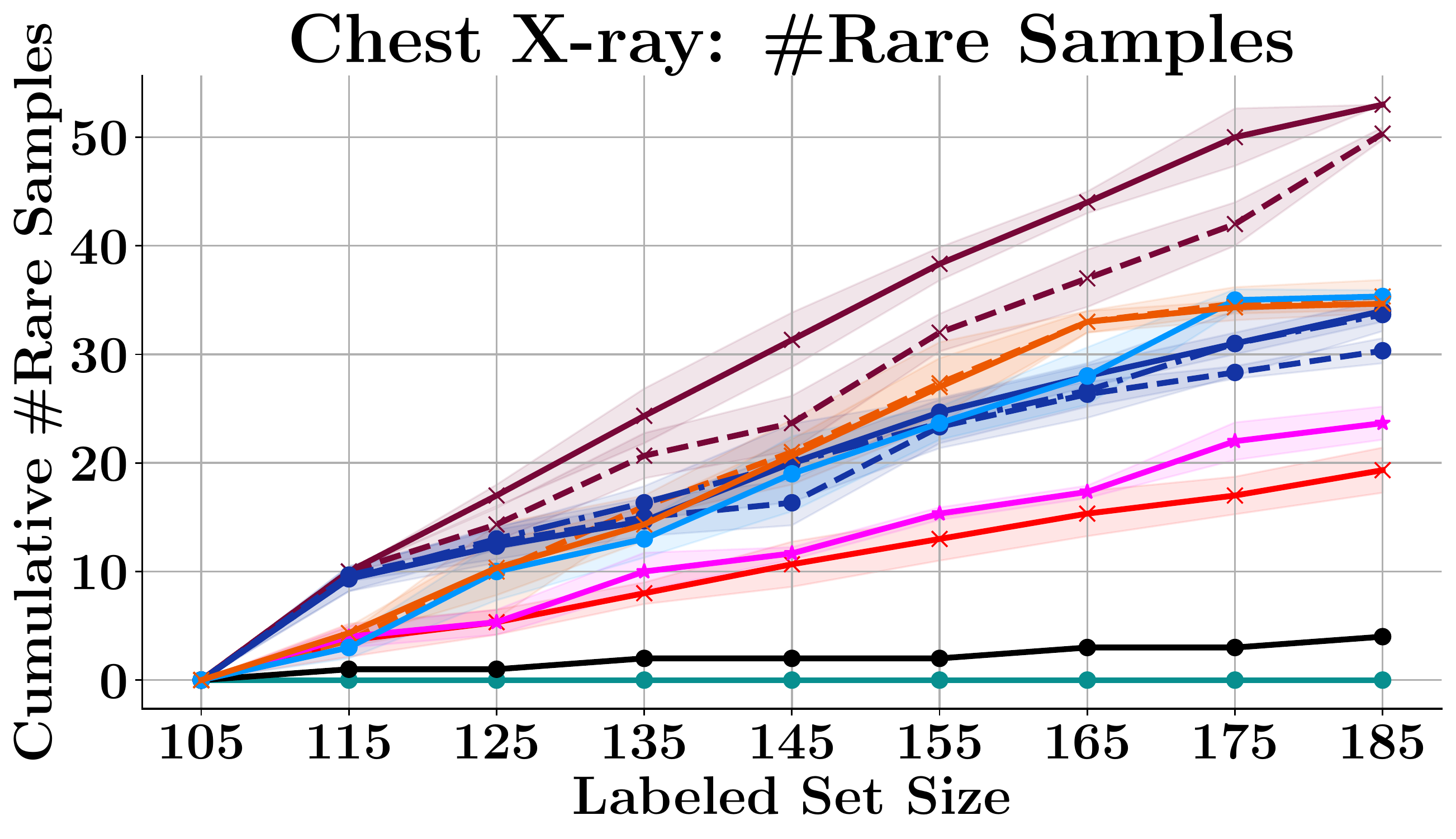}
% \caption{}
\end{subfigure}
\begin{subfigure}[]{0.33\textwidth}
\includegraphics[width = \textwidth]{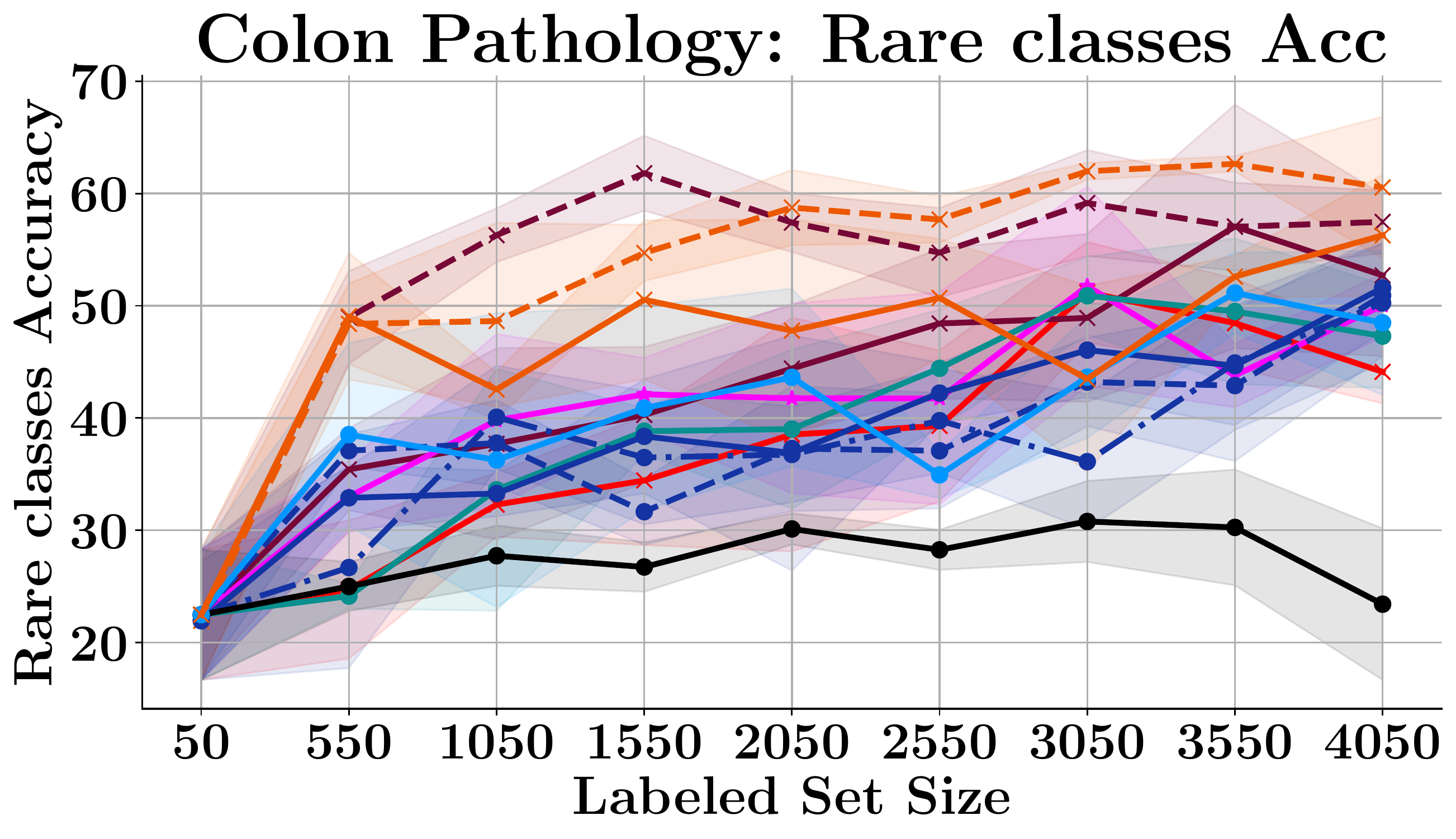}
% \caption{}
\end{subfigure}
\begin{subfigure}[]{0.33\textwidth}
\includegraphics[width = \textwidth]{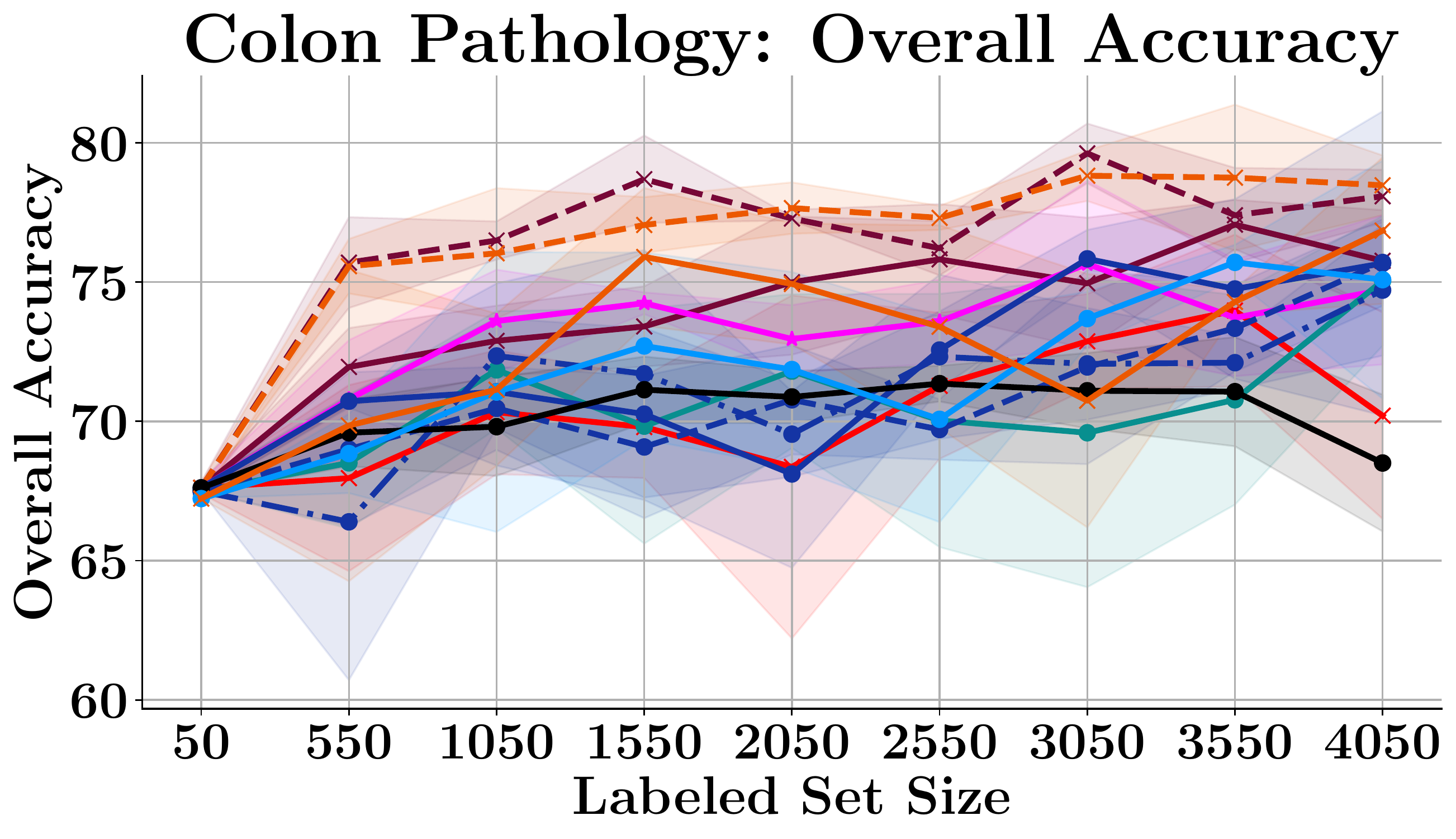}
% \caption{}
\end{subfigure}
\begin{subfigure}[]{0.32\textwidth}
\includegraphics[width = \textwidth]{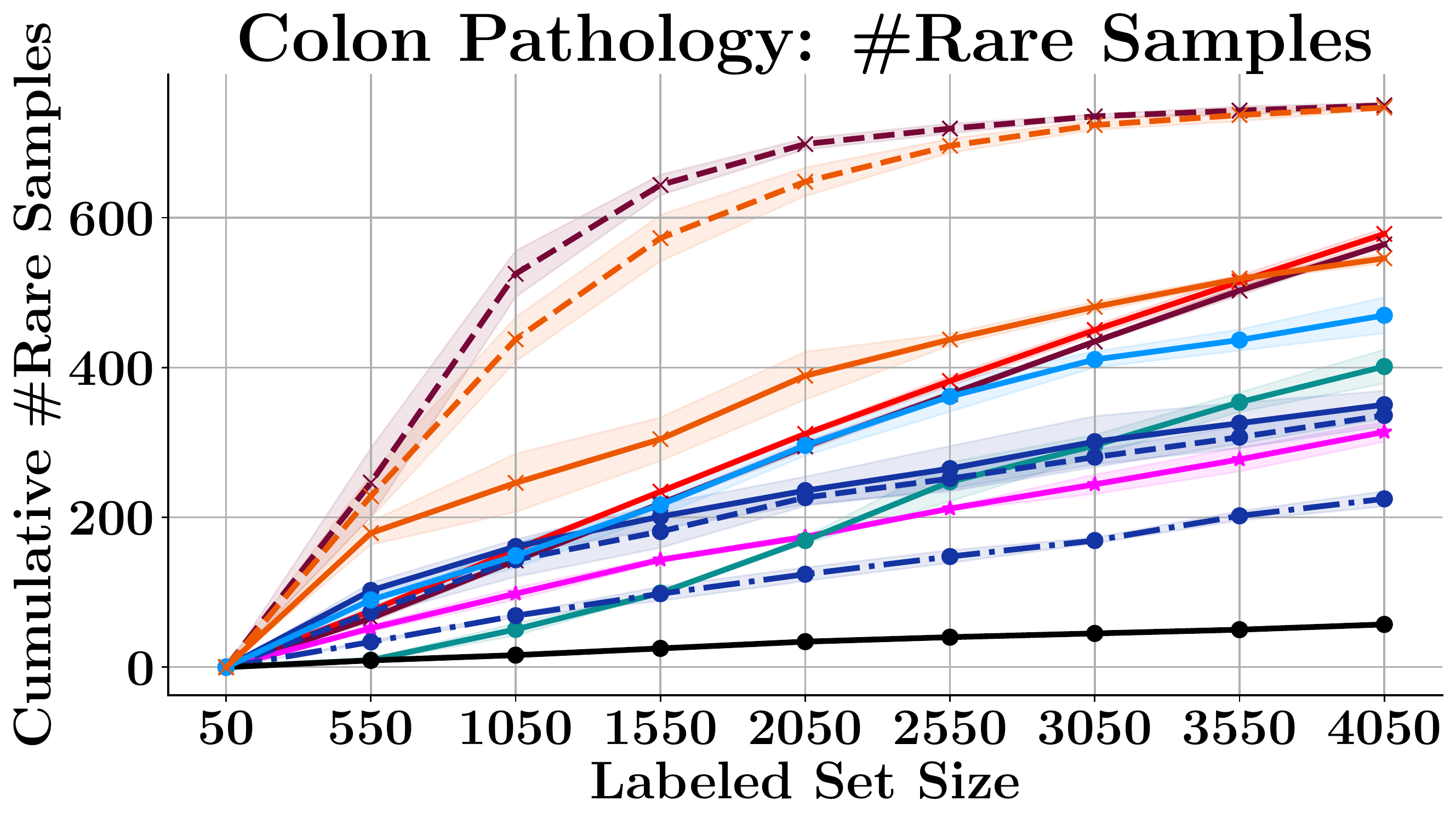}
% \caption{}
\end{subfigure}
\begin{subfigure}[]{0.33\textwidth}
\includegraphics[width = \textwidth]{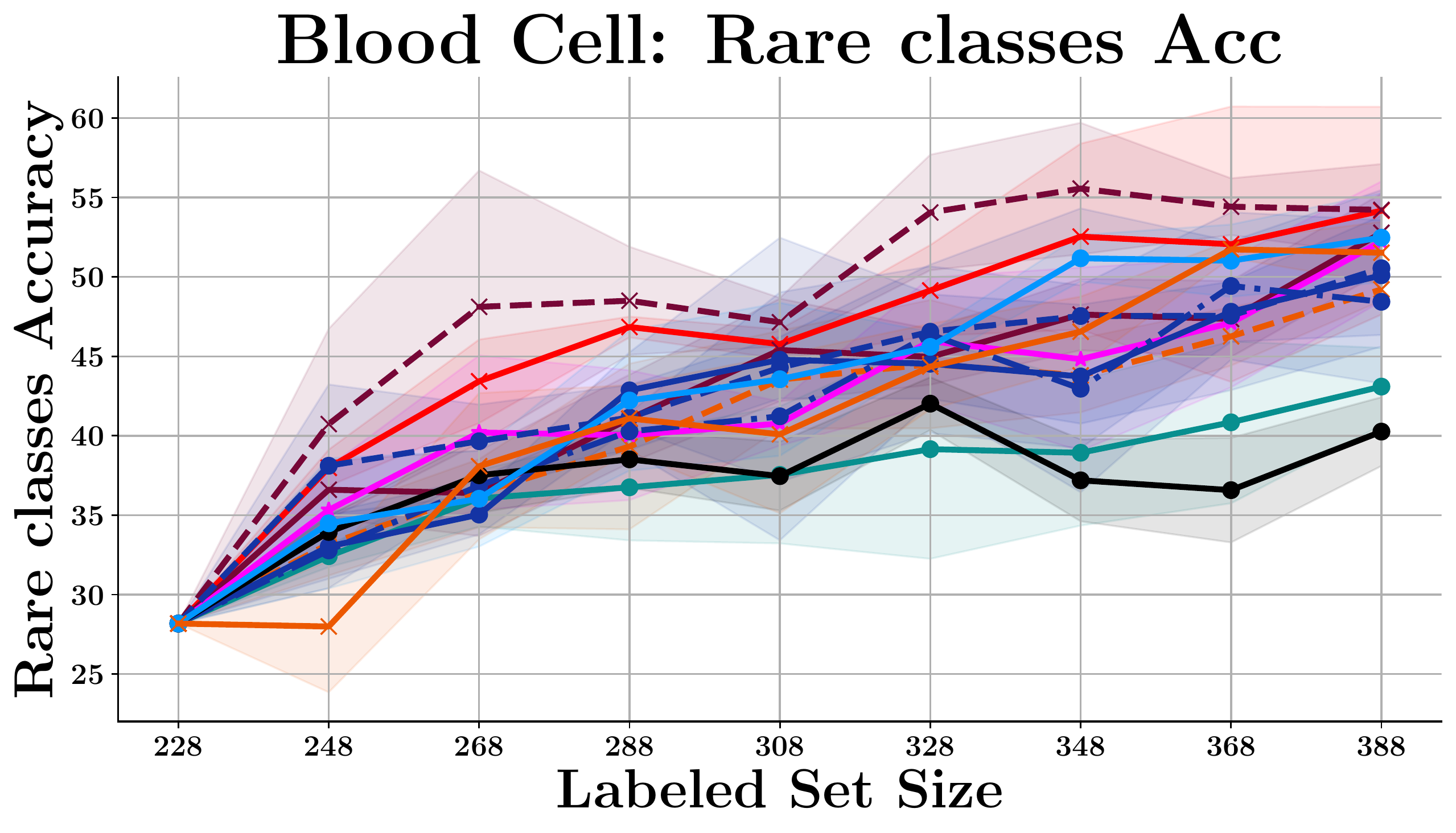}
% \caption{}
\end{subfigure}
\begin{subfigure}[]{0.33\textwidth}
\includegraphics[width = \textwidth]{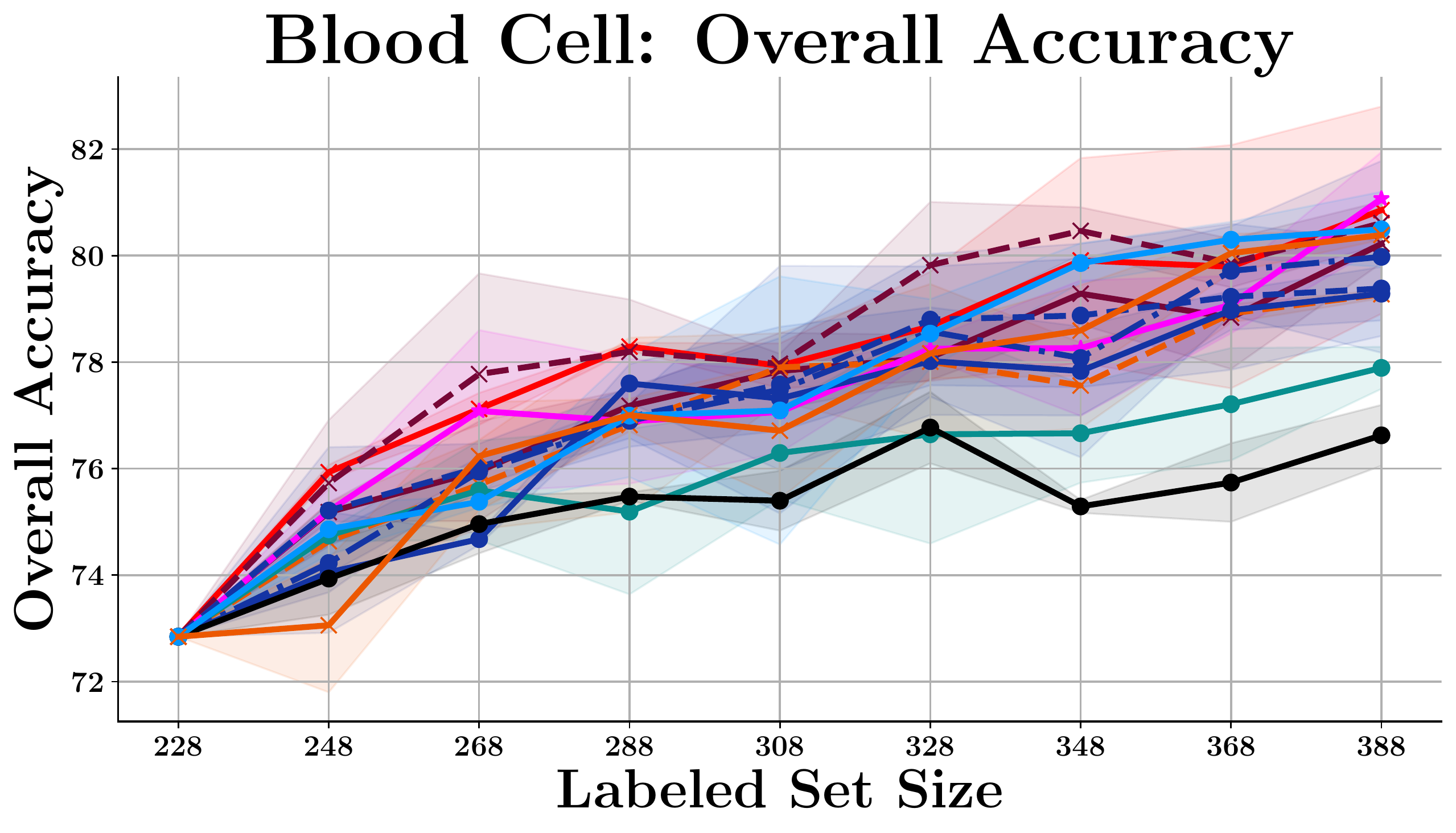}
% \caption{}
\end{subfigure}
\begin{subfigure}[]{0.32\textwidth}
\includegraphics[width = \textwidth]{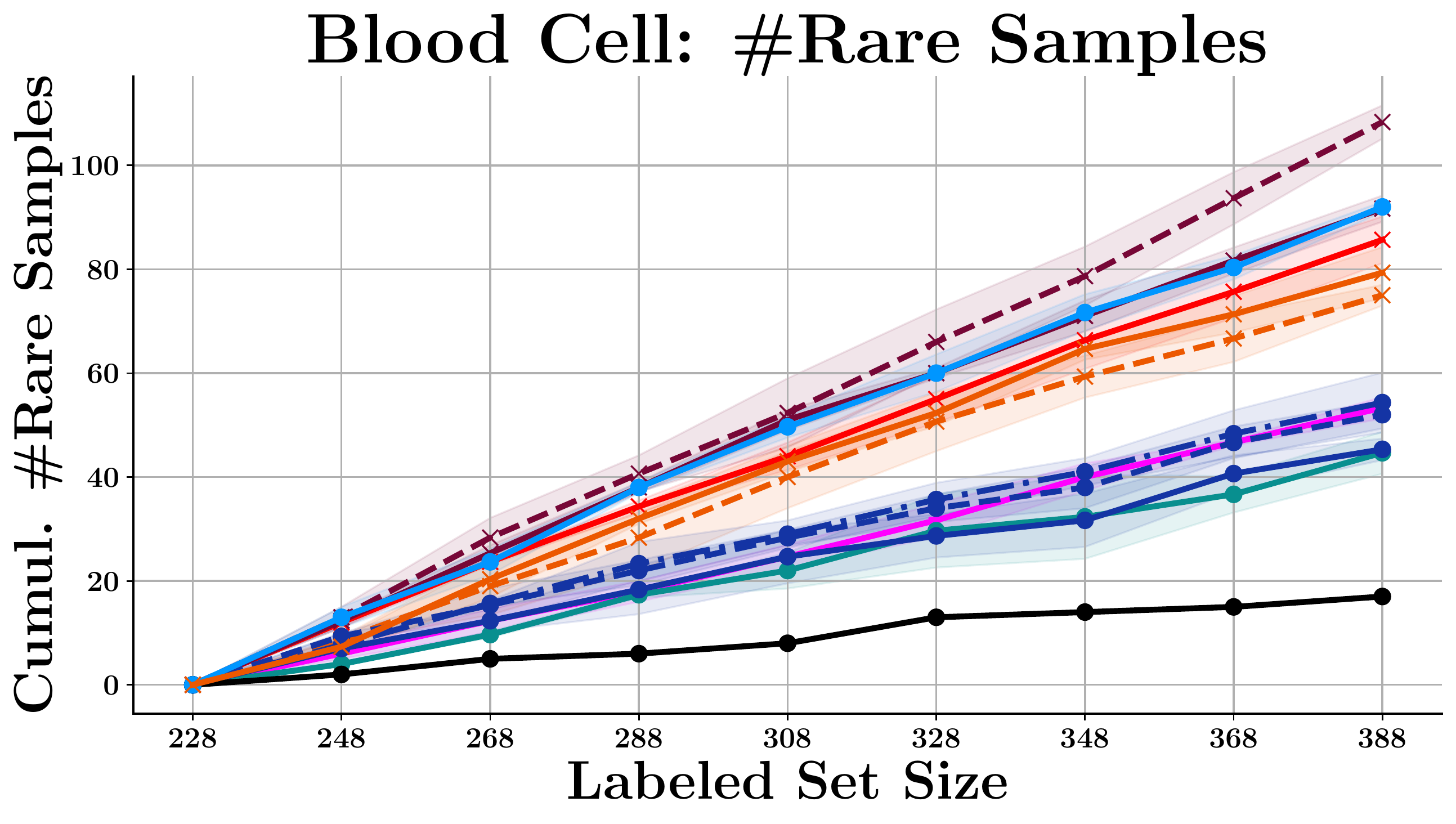}
% \caption{}
\end{subfigure}
\caption{AL for binary imbalanced medical image classification on Pneumonia-MNIST \cite{kermany2018identifying} (\textbf{first} row), Path-MNIST \cite{kather2019predicting} (\textbf{second} row), and Blood-MNIST \cite{acevedo2020dataset} (\textbf{third} row). \model\ outperforms the existing AL methods by $\approx 2\% - 12\%$ on the rare classes acc. (\textbf{left} col.) and $\approx 2\% - 6\%$ on overall acc. (\textbf{center} col.). \textsc{Smi} functions select the most number of rare class samples (\textbf{right} col.)} 
\vspace{-4ex}
% Particularly, \textsc{Flqmi} and \textsc{Flvmi} are the best performing \textsc{Smi} functions.}
\label{fig:res_rare}

\end{figure*}
\vspace{-5ex}

\subsection{Binary Imbalance} \label{sec:exp_binary_imb}

\noindent \textbf{Datasets:}
We apply our framework to \textbf{1)}Pneumonia-MNIST (pediatric chest X-ray)~\cite{medmnistv2, kermany2018identifying}, \textbf{2)}Path-MNIST (colorectal cancer histology)~\cite{medmnistv2, kather2019predicting}, and \textbf{3)}Blood-MNIST (blood cell microscope) \cite{medmnistv2, acevedo2020dataset} medical image classification datasets. To create a more realistic medical scenario, we create a custom dataset that simulates binary class imbalance for each of these datasets for our experiments. Let $\Ccal$ be the set of data points from the rare classes and $\Dcal$ be the set of data points from the over-represented classes. We create the initial labeled set $\Lcal$ (seed set) in AL, $|\Dcal_\Lcal|$ = $\rho |\Ccal_\Lcal|$ and an unlabeled set $\Ucal$ such that $|\Dcal_\Ucal|$ = $\rho |\Ccal_\Ucal|$, where $\rho$ is the imbalance factor. We use a small held out target set $\Tcal$ which contains data points from the rare classes. For Path-MNIST and PneumoniaMNIST, we use $\rho=20$, and for Blood-MNIST, we use $\rho=7$ due to the small size of the dataset. For Pneumonia-MNIST, $|\Ccal_\Lcal| + |\Dcal_\Lcal| = 105$, $|\Ccal_\Ucal| + |\Dcal_\Ucal| = 1100$, $B=10$ (AL batch size) and, $|\Tcal| = 5$. Following the natural class imbalance, we use the `pneumonia' class as the rare class.
For Path-MNIST, $|\Ccal_\Lcal| + |\Dcal_\Lcal| = 3550$, $|\Ccal_\Ucal| + |\Dcal_\Ucal| = 56.8K$, $B=500$ and, $|\Tcal| = 20$. Following the natural class imbalance, we use two classes from the dataset (`mucus', `normal colon mucosa') as rare classes. 
For Blood-MNIST, $|\Ccal_\Lcal| + |\Dcal_\Lcal| = 228$, $|\Ccal_\Ucal| + |\Dcal_\Ucal| = 1824$, $B=20$ and, $|\Tcal| = 20$. Following the natural class imbalance, we use four classes from the dataset (`basophil', `eosinophil', `lymphocyte', `neutrophil') as rare classes.

\noindent\textbf{Results:} The results for the binary imbalance scenario are shown in Fig.~\ref{fig:res_rare}. We observe that the \model\ consistently outperform other methods by $\approx 2\% - 12\%$ on the rare classes accuracy (\figref{fig:res_rare}(left column)) and $\approx 2\% - 6\%$ on overall accuracy (\figref{fig:res_rare}(center column)). This is due to the fact that the \textsc{Smi} functions are able to select significantly more data points that belong to the rare classes (\figref{fig:res_rare}(right column)). Particularly, we observe that when the data modality is \emph{X-ray} (Pneumonia-MNIST), the facility location based \textsc{Smi} variants, \textsc{Flvmi} and \textsc{Flqmi} perform significantly better than other acquistion functions due to their ability to model \emph{representation}. For the \emph{colon pathology} modality (Path-MNIST), \textsc{Gcmi} and \textsc{Flqmi} functions that model \emph{query-relevance} significantly outperform other methods. Lastly, for the blood cell microscope modality (Blood-MNIST), we observe some improvement using \textsc{Flqmi}, although it selects many points from the rare classes. \looseness-1

% Moreover, we observe that the facility location based \textsc{Smi} variants, \textsc{Flvmi} and \textsc{Flqmi}, outperform \textsc{Gcmi} and \textsc{Logdetmi}, since they model diversity and relevance better for this domain. 

\begin{figure*}[h]
\centering
\includegraphics[width = 12cm, height=1cm]{plots/classimb_legend.pdf}
% \centering
\begin{subfigure}[]{0.33\textwidth}
\includegraphics[width = \textwidth]{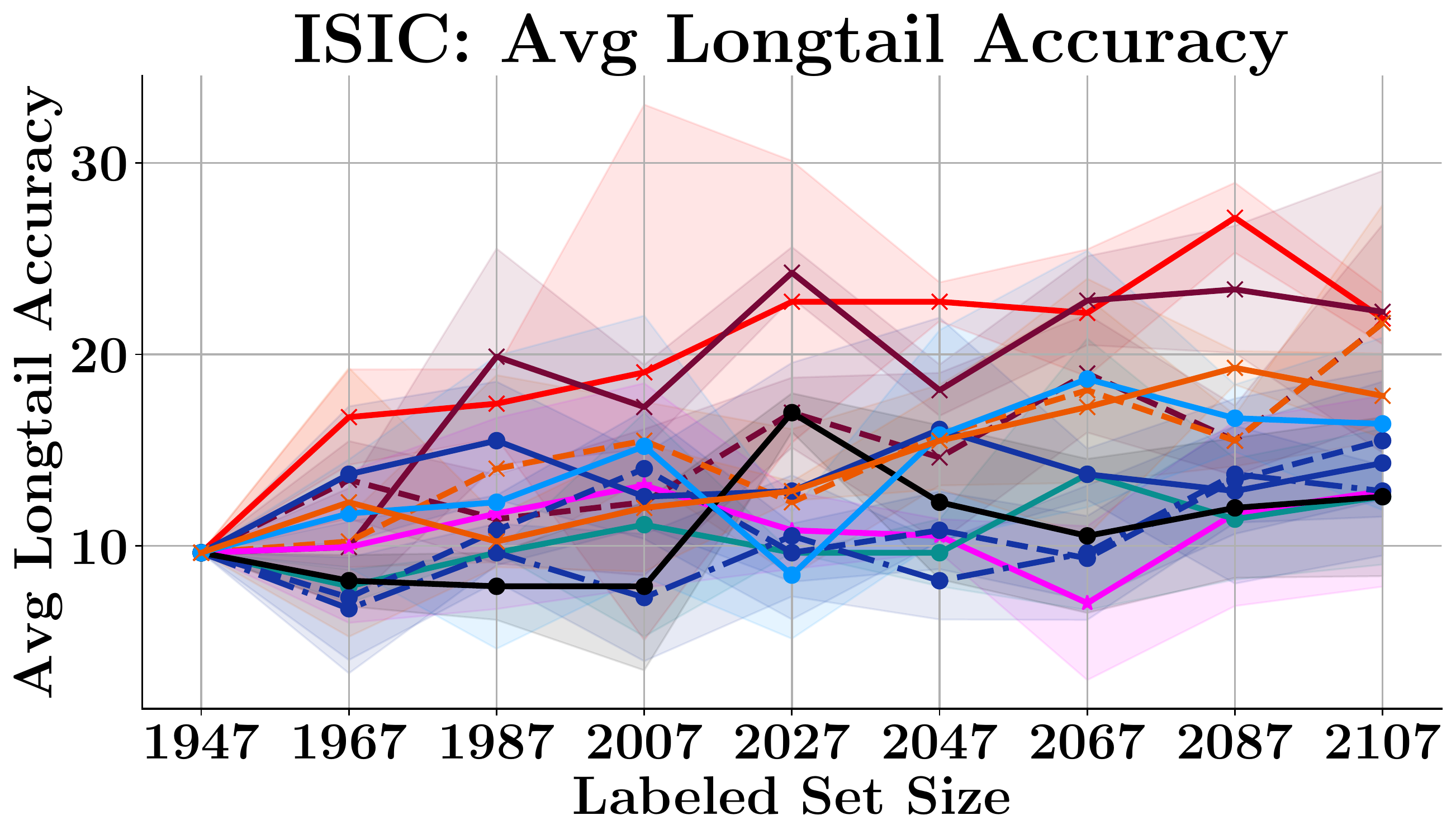}
% \caption{CMI vs Baselines (CIFAR-10)}
% \caption{}
\end{subfigure}
\begin{subfigure}[]{0.33\textwidth}
\includegraphics[width = \textwidth]{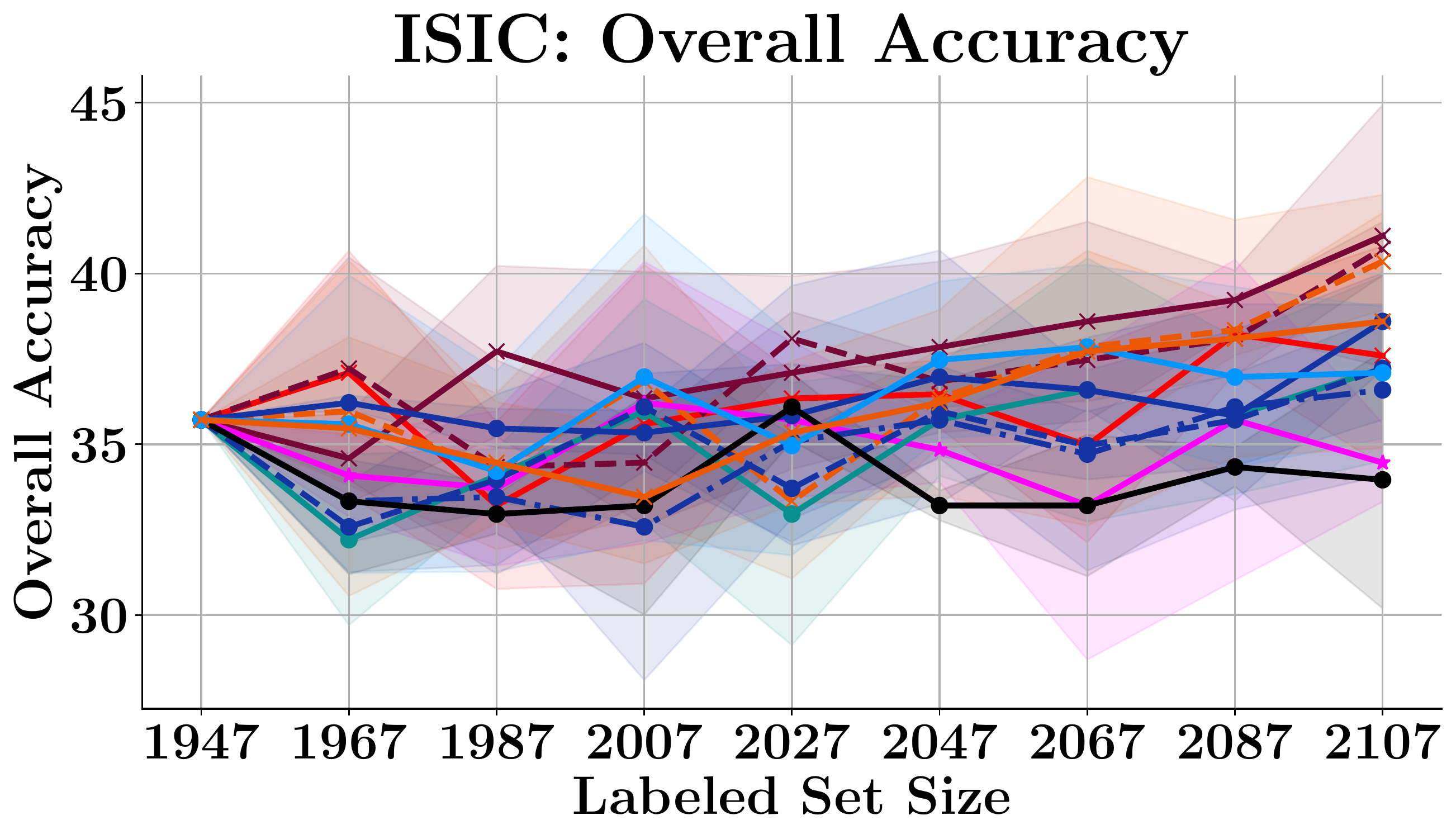}
% \caption{}
\end{subfigure}
\begin{subfigure}[]{0.32\textwidth}
\includegraphics[width = \textwidth]{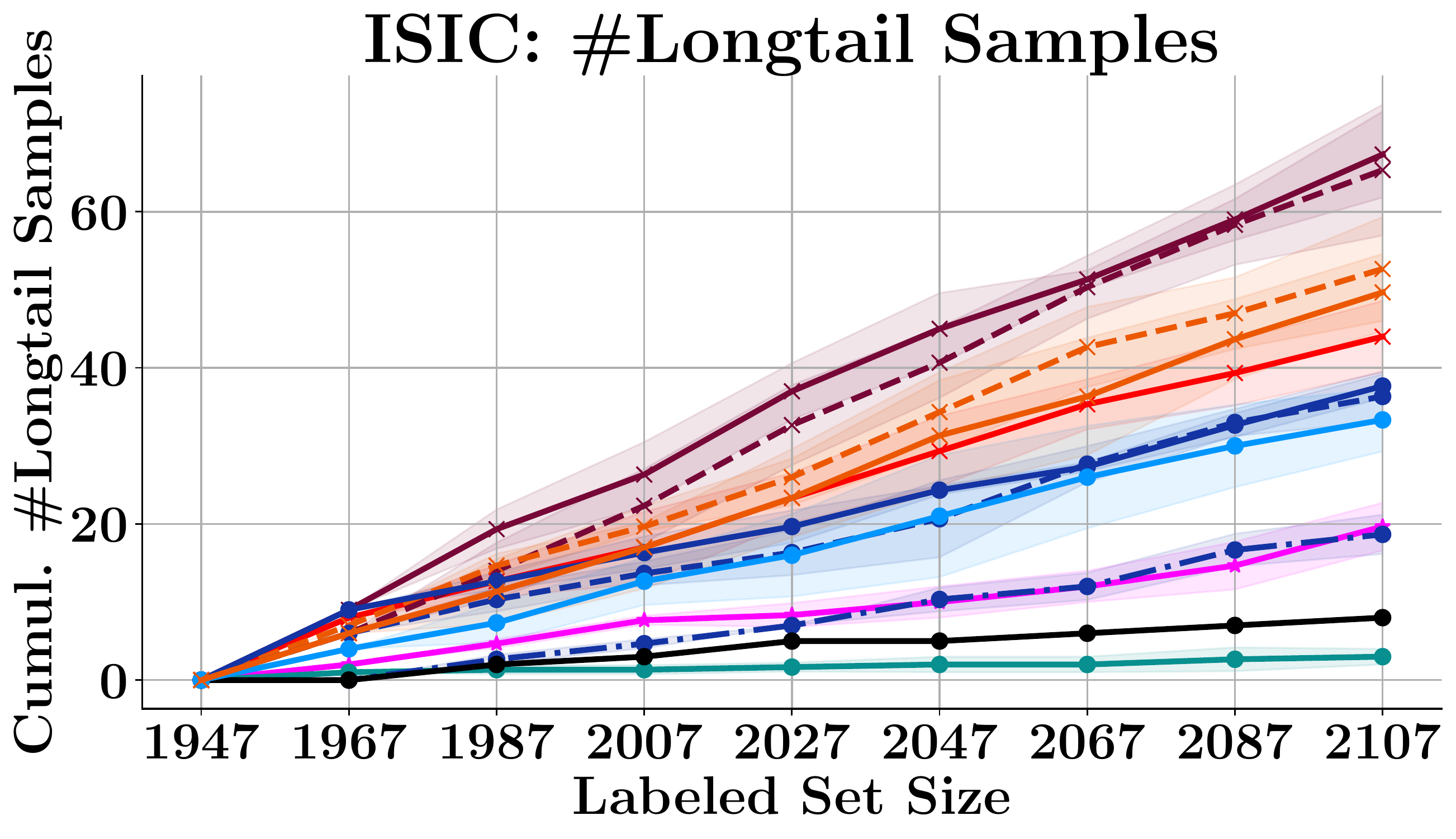}
% \caption{}
\end{subfigure}
\begin{subfigure}[]{0.33\textwidth}
\includegraphics[width = \textwidth]{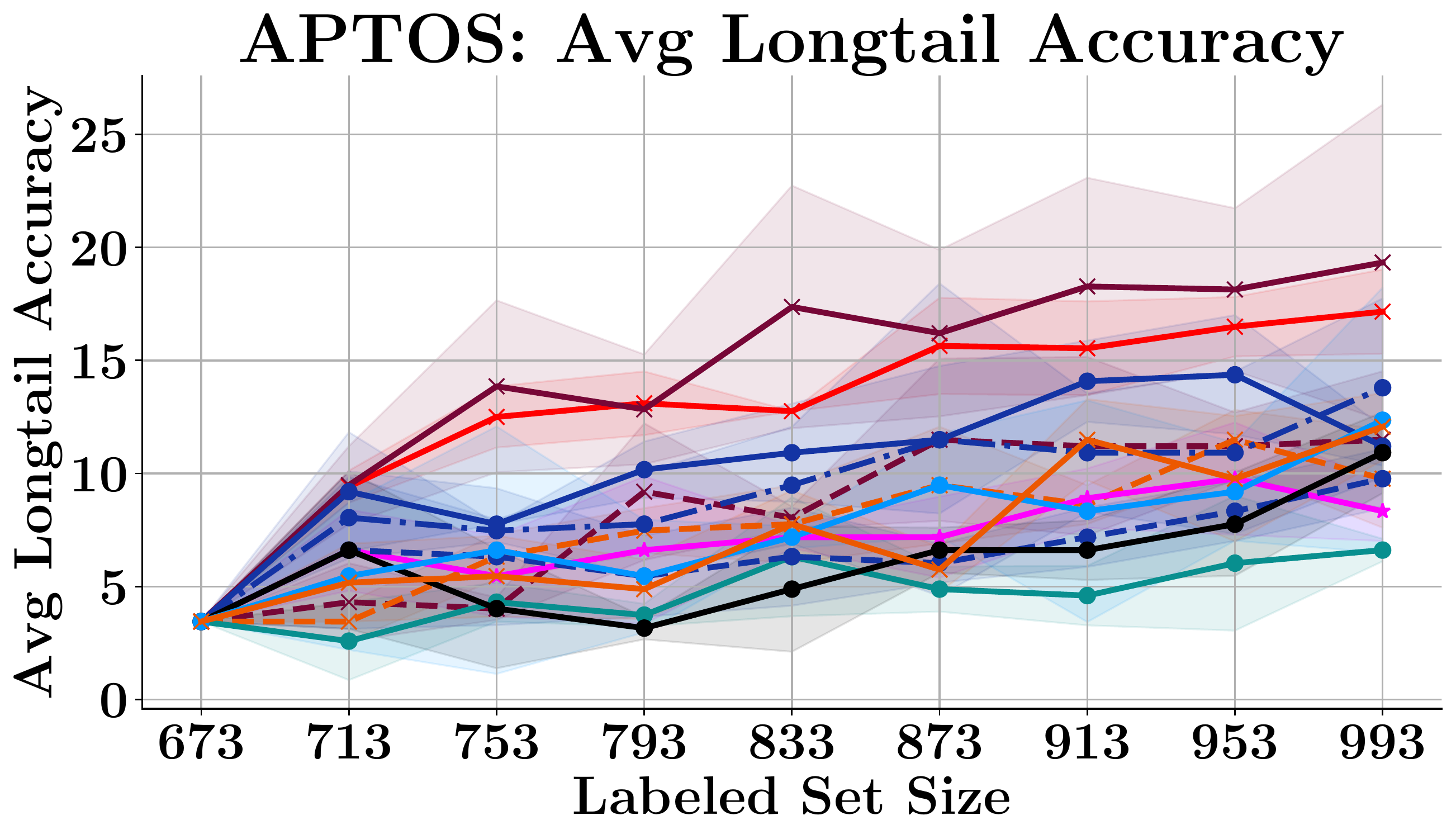}
% \caption{}
\end{subfigure}
\begin{subfigure}[]{0.33\textwidth}
\includegraphics[width = \textwidth]{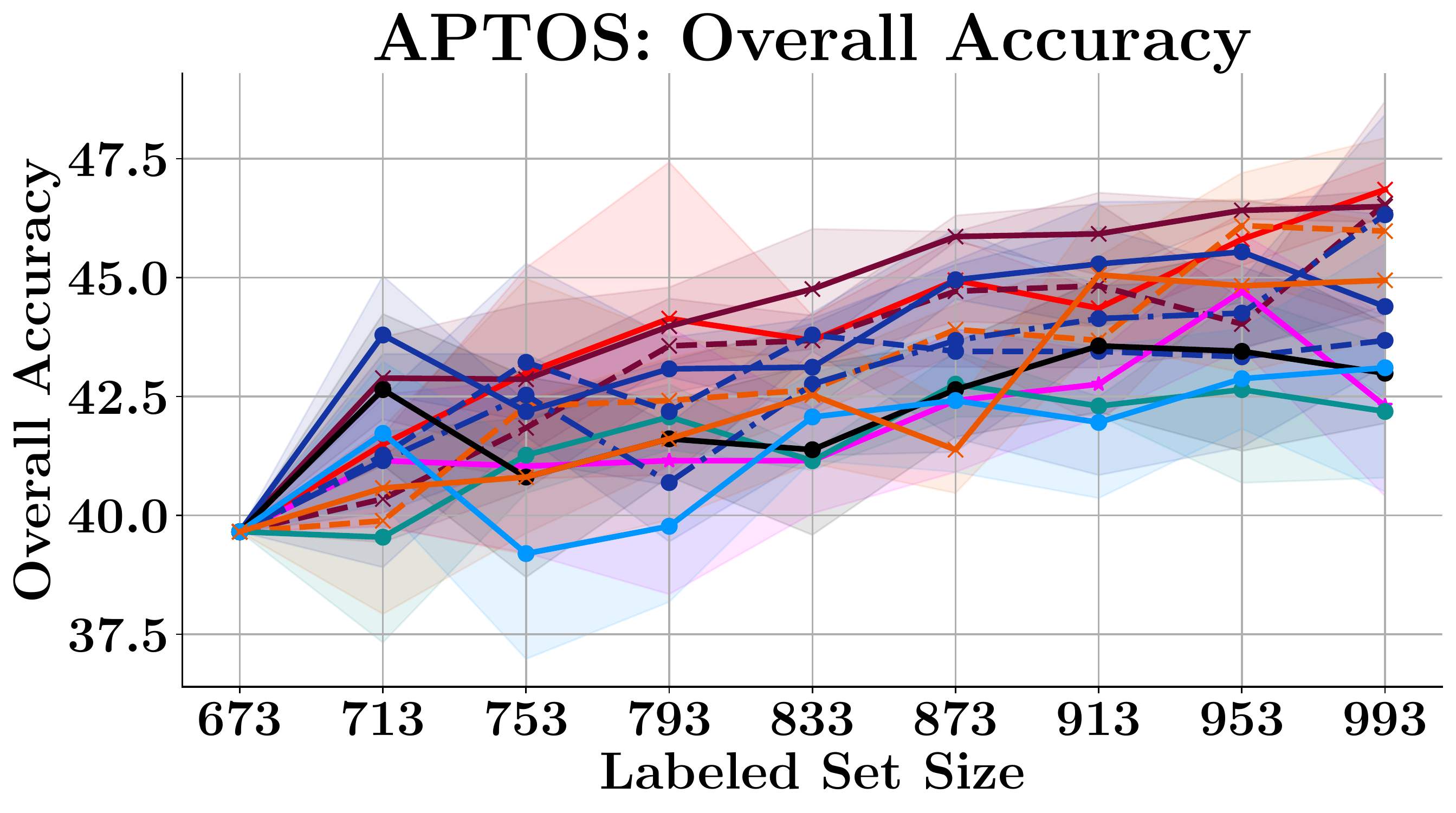}
% \caption{}
\end{subfigure}
\begin{subfigure}[]{0.32\textwidth}
\includegraphics[width = \textwidth]{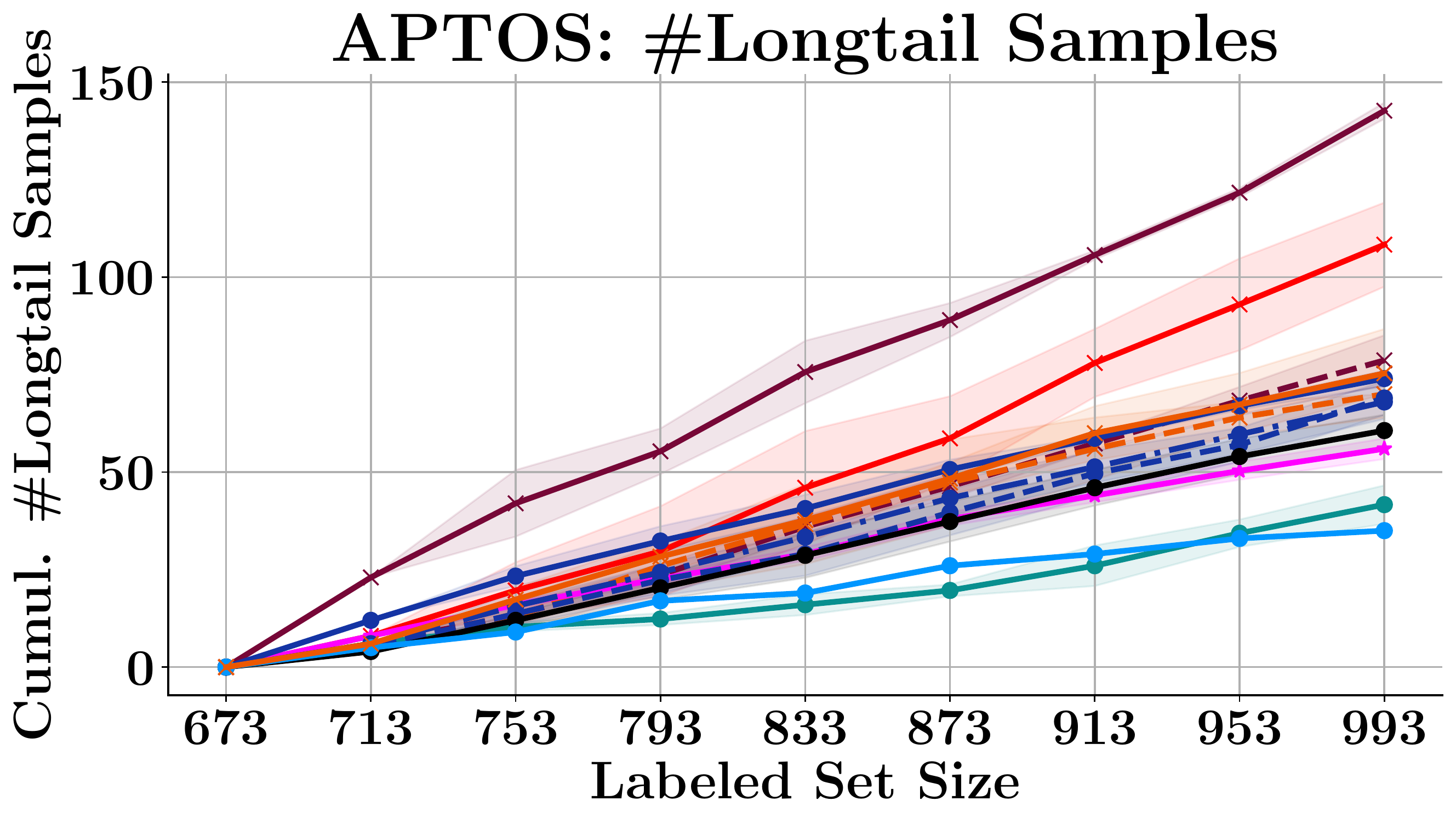}
% \caption{}
\end{subfigure}
\caption{Active learning for long-tail imbalanced medical image classification on ISIC-2018 \cite{codella2019skin} (\textbf{first} row) and APTOS-2019 \cite{aptos2019} (\textbf{second} row). \model\ outperforms the state-of-the-art AL methods by $\approx 10\% - 12\%$ on the average long-tail accuracy (\textbf{left} col.) and $\approx 2\% - 5\%$ on overall accuracy (\textbf{center} col.). \textsc{Smi} functions select the most number of long-tail class samples (\textbf{right} col.)}
\label{fig:res_longtail}
\end{figure*}
\vspace{-2ex}

\subsection{Long-tail Imbalance} \label{sec:exp_longtail_imb}

\noindent \textbf{Datasets:} We apply \model\ to two datasets that naturally show a long-tail distribution: \textbf{1)}The ISIC-2018 skin lesion diagnosis dataset \cite{codella2019skin} and \textbf{2)}APTOS-2019 \cite{aptos2019} for diabetic retinopathy (DR) grading from retinal fundus images. We evaluate all AL methods on a balanced test set to obtain a fair estimate of accuracy across all classes. We split the remaining data randomly with 20\% into the initial labeled set $\Lcal$ and 80\% into the unlabeled set $\Ucal$. We use a small held-out target set $\Tcal$ with data points from the classes at the tail of the distribution (long-tail classes, see \figref{fig:motivating-scenarios}). For ISIC-2018, we use the bottom three infrequent skin lesions from the tail of the distribution as long-tail classes (`bowen's disease', `vascular lesions', and `dermatofibroma'). We set $B=40$ and $|\Tcal|=15$. For APTOS-2019 we use the bottom two infrequent DR gradations as long-tail classes (`severe DR' and `proliferative DR') (see \figref{fig:motivating-scenarios}). We set $B=20$ and $|\Tcal|=10$.
\vspace{0.2cm}

\noindent \textbf{Results:} We present the results for the long-tail imbalance scenario in \figref{fig:res_longtail}. We observe that \model\ consistently outperform other methods by $\approx 10\% - 12\%$ on the average long-tail classes accuracy (\figref{fig:res_longtail}(left column)) and $\approx 2\% - 5\%$ on the overall accuracy (\figref{fig:res_longtail}(center column)). This is because the \textsc{Smi} functions select significantly more data points from the long-tail classes (\figref{fig:res_longtail}(right column)). On both datasets, we observe that the functions modeling query-relevance \emph{and} diversity (\textsc{Flvmi} and \textsc{Logdetmi}) outperform the functions modeling \emph{only} query-relevance (\textsc{Flqmi} and \textsc{Gcmi}). In addition to the ISIC-2018 dataset, we conduct additional class imbalance experiments on the Derma-MNIST \cite{tschandl2018ham10000} dataset (see \Appref{app:dermamnist_res}), and observe that \textsc{Flvmi} and \textsc{Logdetmi} perform significantly better than other functions on the dermatoscopy modality.

% \vspace{-2ex}
\section{Conclusion}
We demonstrate the effectiveness of \model\ for a wide range of medical data modalities for binary and long-tail imbalance. We empirically observe that the current methods in active learning cannot be directly applied to medical datasets with rare classes, and show that a targeting mechanism like \textsc{Smi} can greatly improve the performance on rare classes. 
% A potential negative impact of this work can be the flip side of mitigating class imbalance, \emph{i.e,} the method can also be used to increase class imbalance by targeting to select the over-represented classes.
%%%%%%%%%%%%%%%%%%%%%%%%%%%%%%%%%%%%%%%%%%%%%%%%%%%%%%%%%%%%

\bibliographystyle{splncs04}
\bibliography{main}

\begin{thebibliography}{10}
\providecommand{\url}[1]{\texttt{#1}}
\providecommand{\urlprefix}{URL }
\providecommand{\doi}[1]{https://doi.org/#1}

\bibitem{acevedo2020dataset}
Acevedo, A., Merino, A., Alf{\'e}rez, S., Molina, {\'A}., Bold{\'u}, L.,
  Rodellar, J.: A dataset of microscopic peripheral blood cell images for
  development of automatic recognition systems. Data in Brief, ISSN: 23523409,
  Vol. 30,(2020)  (2020)

\bibitem{kmeansplus}
Arthur, D., Vassilvitskii, S.: k-means++: the advantages of careful seeding.
  In: SODA '07: Proceedings of the eighteenth annual ACM-SIAM symposium on
  Discrete algorithms. pp. 1027--1035. Society for Industrial and Applied
  Mathematics, Philadelphia, PA, USA (2007)

\bibitem{ash2020deep}
Ash, J.T., Zhang, C., Krishnamurthy, A., Langford, J., Agarwal, A.: Deep batch
  active learning by diverse, uncertain gradient lower bounds. In: ICLR (2020)

\bibitem{codella2019skin}
Codella, N., Rotemberg, V., Tschandl, P., Celebi, M.E., Dusza, S., Gutman, D.,
  Helba, B., Kalloo, A., Liopyris, K., Marchetti, M., et~al.: Skin lesion
  analysis toward melanoma detection 2018: A challenge hosted by the
  international skin imaging collaboration (isic). arXiv preprint
  arXiv:1902.03368  (2019)

\bibitem{fujishige2005submodular}
Fujishige, S.: Submodular functions and optimization. Elsevier (2005)

\bibitem{levin2020online}
Gupta, A., Levin, R.: The online submodular cover problem. In: ACM-SIAM
  Symposium on Discrete Algorithms (2020)

\bibitem{he2016deep}
He, K., Zhang, X., Ren, S., Sun, J.: Deep residual learning for image
  recognition. In: Proceedings of the IEEE conference on computer vision and
  pattern recognition. pp. 770--778 (2016)

\bibitem{iyer2020submodular}
Iyer, R., Khargoankar, N., Bilmes, J., Asnani, H.: Submodular combinatorial
  information measures with applications in machine learning. arXiv preprint
  arXiv:2006.15412  (2020)

\bibitem{iyer2015submodular}
Iyer, R.K.: Submodular optimization and machine learning: Theoretical results,
  unifying and scalable algorithms, and applications. Ph.D. thesis (2015)

\bibitem{aptos2019}
Kaggle: Aptos 2019 blindness detection (2019),
  \url{https://www.kaggle.com/c/aptos2019- blindness-detection/data}

\bibitem{kather2019predicting}
Kather, J.N., Krisam, J., Charoentong, P., Luedde, T., Herpel, E., Weis, C.A.,
  Gaiser, T., Marx, A., Valous, N.A., Ferber, D., et~al.: Predicting survival
  from colorectal cancer histology slides using deep learning: A retrospective
  multicenter study. PLoS medicine  \textbf{16}(1),  e1002730 (2019)

\bibitem{kermany2018identifying}
Kermany, D.S., Goldbaum, M., Cai, W., Valentim, C.C., Liang, H., Baxter, S.L.,
  McKeown, A., Yang, G., Wu, X., Yan, F., et~al.: Identifying medical diagnoses
  and treatable diseases by image-based deep learning. Cell  \textbf{172}(5),
  1122--1131 (2018)

\bibitem{killamsetty2021grad}
Killamsetty, K., Durga, S., Ramakrishnan, G., De, A., Iyer, R.: Grad-match:
  Gradient matching based data subset selection for efficient deep model
  training. In: International Conference on Machine Learning. pp. 5464--5474.
  PMLR (2021)

\bibitem{killamsetty2020glister}
Killamsetty, K., Sivasubramanian, D., Ramakrishnan, G., Iyer, R.: Glister:
  Generalization based data subset selection for efficient and robust learning.
  In AAAI  (2021)

\bibitem{kirsch2019batchbald}
Kirsch, A., Van~Amersfoort, J., Gal, Y.: Batchbald: Efficient and diverse batch
  acquisition for deep bayesian active learning. arXiv preprint
  arXiv:1906.08158  (2019)

\bibitem{kothawade2021similar}
Kothawade, S., Beck, N., Killamsetty, K., Iyer, R.: Similar: Submodular
  information measures based active learning in realistic scenarios. Advances
  in Neural Information Processing Systems  \textbf{34} (2021)

\bibitem{kothawade2021talisman}
Kothawade, S., Ghosh, S., Shekhar, S., Xiang, Y., Iyer, R.: Talisman: Targeted
  active learning for object detection with rare classes and slices using
  submodular mutual information. arXiv preprint arXiv:2112.00166  (2021)

\bibitem{kothawade2021prism}
Kothawade, S., Kaushal, V., Ramakrishnan, G., Bilmes, J., Iyer, R.: Prism: A
  rich class of parameterized submodular information measures for guided subset
  selection. arXiv preprint arXiv:2103.00128  (2021)

\bibitem{kothyari2021personalizing}
Kothyari, M., Mekala, A.R., Iyer, R., Ramakrishnan, G., Jyothi, P.:
  Personalizing asr with limited data using targeted subset selection. arXiv
  preprint arXiv:2110.04908  (2021)

\bibitem{li2012multi}
Li, J., Li, L., Li, T.: Multi-document summarization via submodularity. Applied
  Intelligence  \textbf{37}(3),  420--430 (2012)

\bibitem{lin2012submodularity}
Lin, H.: Submodularity in natural language processing: algorithms and
  applications. Ph.D. thesis (2012)

\bibitem{loshchilov2016sgdr}
Loshchilov, I., Hutter, F.: Sgdr: Stochastic gradient descent with warm
  restarts. arXiv preprint arXiv:1608.03983  (2016)

\bibitem{mirzasoleiman2015lazier}
Mirzasoleiman, B., Badanidiyuru, A., Karbasi, A., Vondr{\'a}k, J., Krause, A.:
  Lazier than lazy greedy. In: Proceedings of the AAAI Conference on Artificial
  Intelligence. vol.~29 (2015)

\bibitem{roth2006margin}
Roth, D., Small, K.: Margin-based active learning for structured output spaces.
  In: European Conference on Machine Learning. pp. 413--424. Springer (2006)

\bibitem{sener2018active}
Sener, O., Savarese, S.: Active learning for convolutional neural networks: A
  core-set approach. In: International Conference on Learning Representations
  (2018)

\bibitem{settles2009active}
Settles, B.: Active learning literature survey. Tech. rep., University of
  Wisconsin-Madison Department of Computer Sciences (2009)

\bibitem{tschandl2018ham10000}
Tschandl, P., Rosendahl, C., Kittler, H.: The ham10000 dataset, a large
  collection of multi-source dermatoscopic images of common pigmented skin
  lesions. Scientific data  \textbf{5}(1), ~1--9 (2018)

\bibitem{vasudevan2017query}
Vasudevan, A.B., Gygli, M., Volokitin, A., Van~Gool, L.: Query-adaptive video
  summarization via quality-aware relevance estimation. In: Proceedings of the
  25th ACM international conference on Multimedia. pp. 582--590 (2017)

\bibitem{wang2014new}
Wang, D., Shang, Y.: A new active labeling method for deep learning. In: 2014
  International joint conference on neural networks (IJCNN). pp. 112--119. IEEE
  (2014)

\bibitem{medmnistv2}
Yang, J., Shi, R., Wei, D., Liu, Z., Zhao, L., Ke, B., Pfister, H., Ni, B.:
  Medmnist v2: A large-scale lightweight benchmark for 2d and 3d biomedical
  image classification. arXiv preprint arXiv:2008  (2021)

\end{thebibliography}

\newpage

\appendix

\setcounter{page}{1}

\section*{Supplementary Material for Targeted Active Learning for Imbalanced Medical Image Classification} 

\section{Summary of Notations}\label{app:notation-summary}

%BIG TABLE STARTS HERE
 \begin{table*}[!h]
 \centering
%  \arrayrulecolor[rgb]{0.192,0.192,0.192}
 %\begin{adjustbox}{max width=0.48\textwidth}
 \begin{tabular}{|l|l|p{0.5\textwidth}|} 
 \toprule
 %\hline 
 \hline 
 \multicolumn{1}{|l|}{Topic} & Notation & Explanation \\ \hline
 \toprule \hline
 &  $\Ucal$ & Unlabeled set of $|\Ucal|$ instances\\ 
 \multicolumn{1}{|p{0.20\textwidth}|}{ \model\ (\secref{sec:our_method})} 
 & $\Acal$ & A subset of $\Ucal$\\ 
 & $s_{ij}$ & Similarity between any two data points $i$ and $j$\\
 & $f$ & A submodular function\\
 & $\Lcal$ & Labeled set of data points\\
 & $\Qcal$ & Query set\\
 & $\Mcal$ & Deep model\\
 & $B$ & Active learning selection budget\\
 & $N$ & Number of selection rounds in active learning\\
 & $\Tcal$ & Held-out target set containing data points from the rare classes\\
 & $\hat{\Tcal}$ & Subset of $\Tcal$ containing only the misclassified data points\\
 & $\Hcal$ & Loss function used to train model $\Mcal$\\
 & $X$ & Pairwise similarity matrix computed using gradients\\
  \multicolumn{1}{|p{0.20\textwidth}|}{ Experiments (\secref{sec:experiments})} 
 & $\Ccal_\Lcal$ & Rare classes data points in the labeled set $\Lcal$\\
 & $\Ccal_\Ucal$ & Rare classes data points in the unlabeled set $\Ucal$\\
 & $\Dcal_\Lcal$ & Non-rare (frequent) classes data points in the labeled set $\Lcal$\\
 & $\Dcal_\Ucal$  & Non-rare (frequent) classes data points in the unlabeled set $\Ucal$\\
 
  \hline
 \bottomrule
 \end{tabular}
 %\end{adjustbox}
%  \arrayrulecolor{black}
 \caption{Summary of notations used throughout this paper}
 \label{tab:main-notations}
 \end{table*}

\section{Details of Datasets used} \label{app:dataset_details}

\subsection{ISIC \cite{codella2019skin}}

\begin{itemize}
    \item ISIC dataset is a representative collection of all important diagnostic categories in the realm of pigmented lesions
    \item It contains, 10015 28x28 color images in 7 different classes 
    \item Classes represent various types of skin cancer diseases like Actinic keratoses and intraepithelial carcinoma / Bowen's disease (akiec), basal cell carcinoma (bcc), benign keratosis-like lesions (solar lentigines / seborrheic keratoses and lichen-planus like keratoses, bkl), dermatofibroma (df), melanoma (mel), melanocytic nevi (nv) and vascular lesions (angiomas, angiokeratomas, pyogenic granulomas and hemorrhage, vasc) 
    % \item Number of images in classes follow longtail distribution
    % \item Split configuration used for this dataset is as follows
    
    % \begin{center}
    %     \begin{tabular}{|c|c|c|c|c|c|c|c|}
    %         \hline
    %         Classes &  0  &  1  &  2  &  3 &  4   &  5   &  6  \\
    %         \hline
    %         Train   & 57  & 95  & 212 & 15  & 215 & 1333 & 20  \\
    %         Query   & 10  & 0   & 0   & 10  & 0   & 0    & 10  \\
    %         Lake    & 222 & 381 & 849 & 52  & 860 & 5334 & 74  \\
    %         Test    & 38  & 38  & 38  & 38  & 38  & 38   & 38  \\
    %         \hline
    %     \end{tabular}
    % \end{center}
    
\end{itemize}

\subsection{APTOS \cite{aptos2019}}

\begin{itemize}
    \item APTOS dataset contains retina images taken using fundus photography under a variety of imaging conditions
    \item It contains 3662 28x28 color images in 5 different classes 
    \item Classes represent severity of diabetic retinopathy on a scale of 0 to 4
    % \item Number of images in classes follow longtail imbalanced distribution    
    % \item Split configuration used for this dataset is as follows
    
    % \begin{center}
    %     \begin{tabular}{|c|c|c|c|c|c|}
    %         \hline
    %         Classes & 0    & 1   & 2    & 3   & 4   \\
    %         \hline
    %         Train   & 349  & 62  & 188  & 27  & 47  \\
    %         Query   & 0    & 0   & 0    & 20  & 20  \\
    %         Lake    & 1398 & 250 & 753  & 88  & 170 \\
    %         Test    & 58   & 58  & 58   & 58  & 58  \\
    %         \hline
    %     \end{tabular}
    % \end{center}
    
\end{itemize}

\subsection{PathMNIST \cite{kather2019predicting}}

\begin{itemize}
    \item A dataset based on a prior study for predicting survival from colorectal cancer histology slides, which provides a dataset NCT-CRC-HE-100K of 100,000 non-overlapping image patches from hematoxylin and eosin stained histological images, and a test dataset CRC-VAL-HE-7K of 7,180 image patches from a different clinical center.
    \item There are 9 types of tissues are involved, resulting in a multi-class classification task.
    \item The images are resized from 3 x 224 x 224 into 3 x 28 x 28 as in \cite{medmnistv2}.
    \item Classes represent 9 types of tissues: 'adipose', 'background', 'debris', 'lymphocytes', 'mucus', 'smooth muscle', 'normal colon mucosa', 'cancer-associated stroma', 'colorectal adenocarcinoma epithelium'.
\end{itemize}

\subsection{PneumoniaMNIST \cite{kermany2018identifying}}

\begin{itemize}
    \item A dataset based on a prior dataset of 5,856 pediatric chest X-ray images. 
    \item The task is binary-class classification of pneumonia and normal. 
    \item We split the source training set with a ratio of 9:1 into training and validation set, and use its source validation set as the test set. 
    \item The source images are single-channel, and their sizes range from (384-2,916) x (127-2,713). We use data from \cite{medmnistv2} where they center-crop the images and resize them into 1 x 28 x 28. 
\end{itemize}

\subsection{BloodMNIST \cite{acevedo2020dataset}}

\begin{itemize}
    \item BloodMNIST dataset is based on a dataset of individual normal blood cells, captured from individuals without any kind of infection or disease
    \item It contains 17092 32x32 color images in 8 different classes 
    \item Classes represent various types of blood cells (without any infection) like basophil, eosinophil, erythroblast, ig, lymphocyte, monocyte, neutrophil, platelet 
    % \item Number of images in classes follow binary imbalanced distribution
    % \item Split configuration used for this dataset is as follows
    
    % \begin{center}
    %     \begin{tabular}{|c|c|c|c|c|c|c|c|c|c|}
    %         \hline
    %         Classes & 0    & 1   & 2   & 3   & 4   & 5   & 6    & 7   \\
    %         \hline
    %         Train   & 7    & 50  & 7   & 50  & 7   & 7   & 50   & 50  \\
    %         Query   & 20   & 0   & 20  & 0   & 20  & 20  & 0    & 0   \\
    %         Lake    & 56   & 400 & 56  & 400 & 56  & 56  & 400  & 400 \\
    %         Test    & 243  & 243 & 243 & 243 & 243 & 243 & 243  & 243 \\
    %         \hline
    %     \end{tabular}
    % \end{center}
    
\end{itemize}

\subsection{DermaMNIST \cite{codella2019skin}}

\begin{itemize}
    \item DermaMNIST dataset is a large collection of multi-source dermatoscopic images of common pigmented skin lesions
    \item It contains, 10015 32x32 color images in 7 different classes.
    \item Classes represent various types of skin lesions like actinic keratoses, basal cell carcinoma, benign keratosis-like lesions, dermatofibroma, melanoma, melanocytic nevi, vascular lesions
    % \item Number of images in classes follow binary imbalanced distribution
    % \item Split configuration used for this dataset is as follows
    
    % \begin{center}
    %     \begin{tabular}{|c|c|c|c|c|c|c|c|}
    %         \hline
    %         Classes &  0  &  1  &  2  &  3 &  4  &  5  &  6  \\
    %         \hline
    %         Train   & 50  & 50  & 50  & 6  & 50  & 50  & 6   \\
    %         Query   & 0   & 0   & 0   & 10 & 0   & 0   & 10  \\
    %         Lake    & 200 & 200 & 200 & 18 & 200 & 200 & 18  \\
    %         Test    & 29  & 29  & 29  & 29 & 29  & 29  & 29  \\
    %         \hline
    %     \end{tabular}
    % \end{center}
    
\end{itemize}

\section{Scalability of \model} \label{app:scalability}

Below, we provide a detailed analysis of the complexity of creating and optimizing the different SMI functions. Denote $|\Xcal|$ as the size of set $\Xcal$. Also, let $|\Ucal| = n$ (the ground set size, which is the size of the unlabeled set in this case). 
\begin{itemize}
    \item \textbf{Facility Location: } We start with FLVMI. The complexity of creating the kernel matrix is $O(n^2)$. The complexity of optimizing it is $\tilde{O}(n^2)$ (using memoization)\footnote{$\tilde{O}$: Ignoring log-factors} if we use the stochastic greedy algorithm~\cite{mirzasoleiman2015lazier} and $O(n^2k)$ with the naive greedy algorithm. The overall complexity is $\tilde{O}(n^2)$.
    For FLQMI, the cost of creating the kernel matrix is $O(n|\Qcal|)$, and the cost of optimization is also $\tilde{O}(n|\Qcal|)$ (with naive greedy, it is $O(nB |\Qcal|)$). 
    \item \textbf{Log-Determinant: } For LogDetMI, the complexity of the kernel matrix computation (and storage) is $O(n^2)$. The complexity of optimizing the LogDet function using the stochastic greedy algorithm is $\tilde{O}(B^2 n)$, so the overall complexity is $\tilde{O}(n^2 + B^2n)$.
    \item \textbf{Graph-Cut: } For GCMI, we require a $O(n|\Qcal|)$ kernel matrix, and the complexity of the stochastic greedy algorithm is also $\tilde{O}(n|\Qcal|)$. 
\end{itemize}
We end with a few comments. First, most of the complexity analysis above is with the stochastic greedy algorithm~\cite{mirzasoleiman2015lazier}. If we use the naive or lazy greedy algorithm, the worst-case complexity is a factor $B$ larger. Secondly, we ignore log-factors in the complexity of stochastic greedy since the complexity is actually $O(n\log 1/\epsilon)$, which achieves an $1 - 1/e - \epsilon$ approximation.

\section{Additional Results} \label{app:add_res}

\subsection{Dermatascope Binary Imbalance Results} \label{app:dermamnist_res}

\begin{figure*}[h!]
\centering
\includegraphics[width = 12cm, height=1cm]{plots/classimb_legend.pdf}
% \centering
\begin{subfigure}[]{0.49\textwidth}
\includegraphics[width = \textwidth]{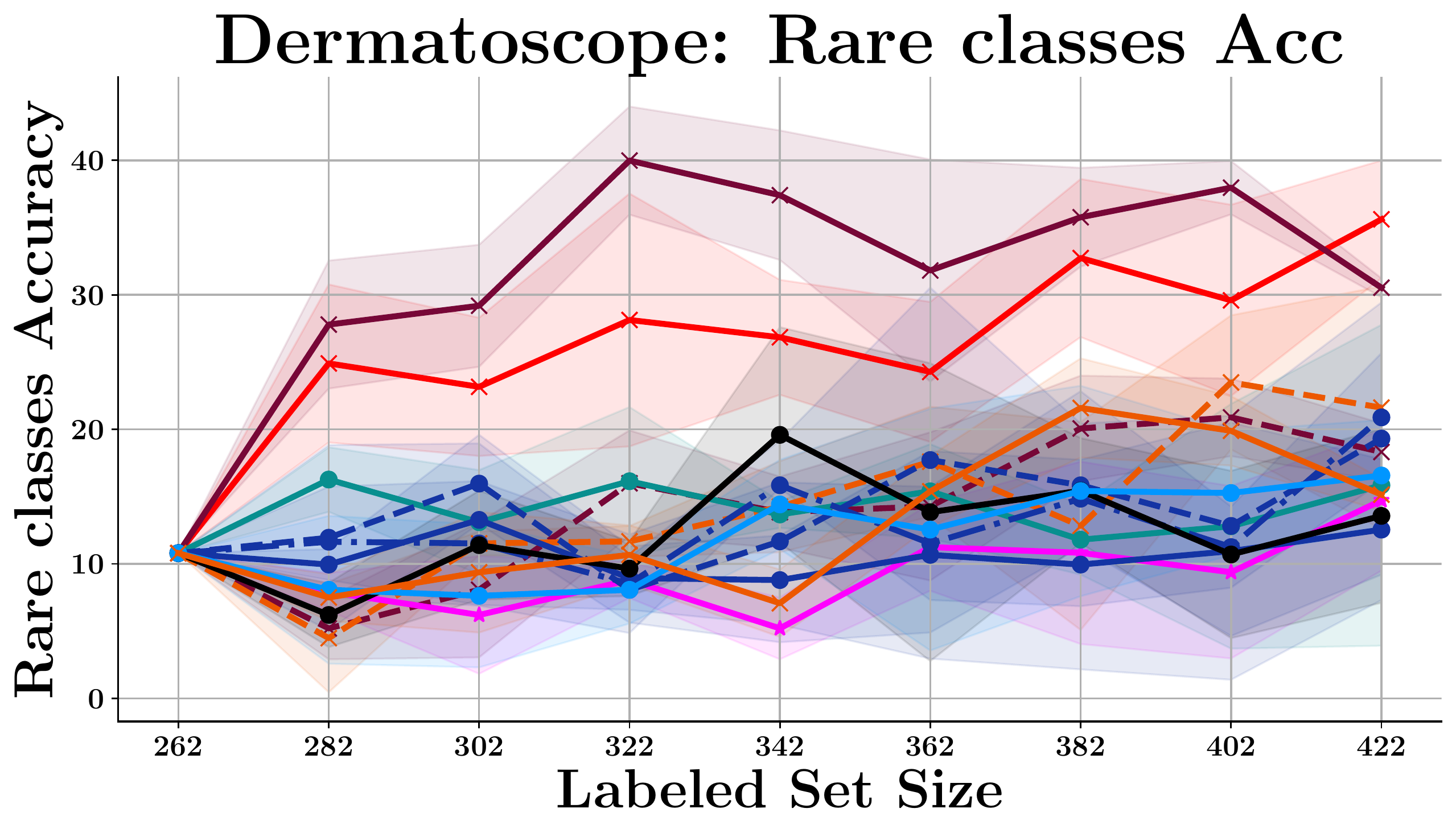}
% \caption{CMI vs Baselines (CIFAR-10)}
% \caption{}
\end{subfigure}
\begin{subfigure}[]{0.49\textwidth}
\includegraphics[width = \textwidth]{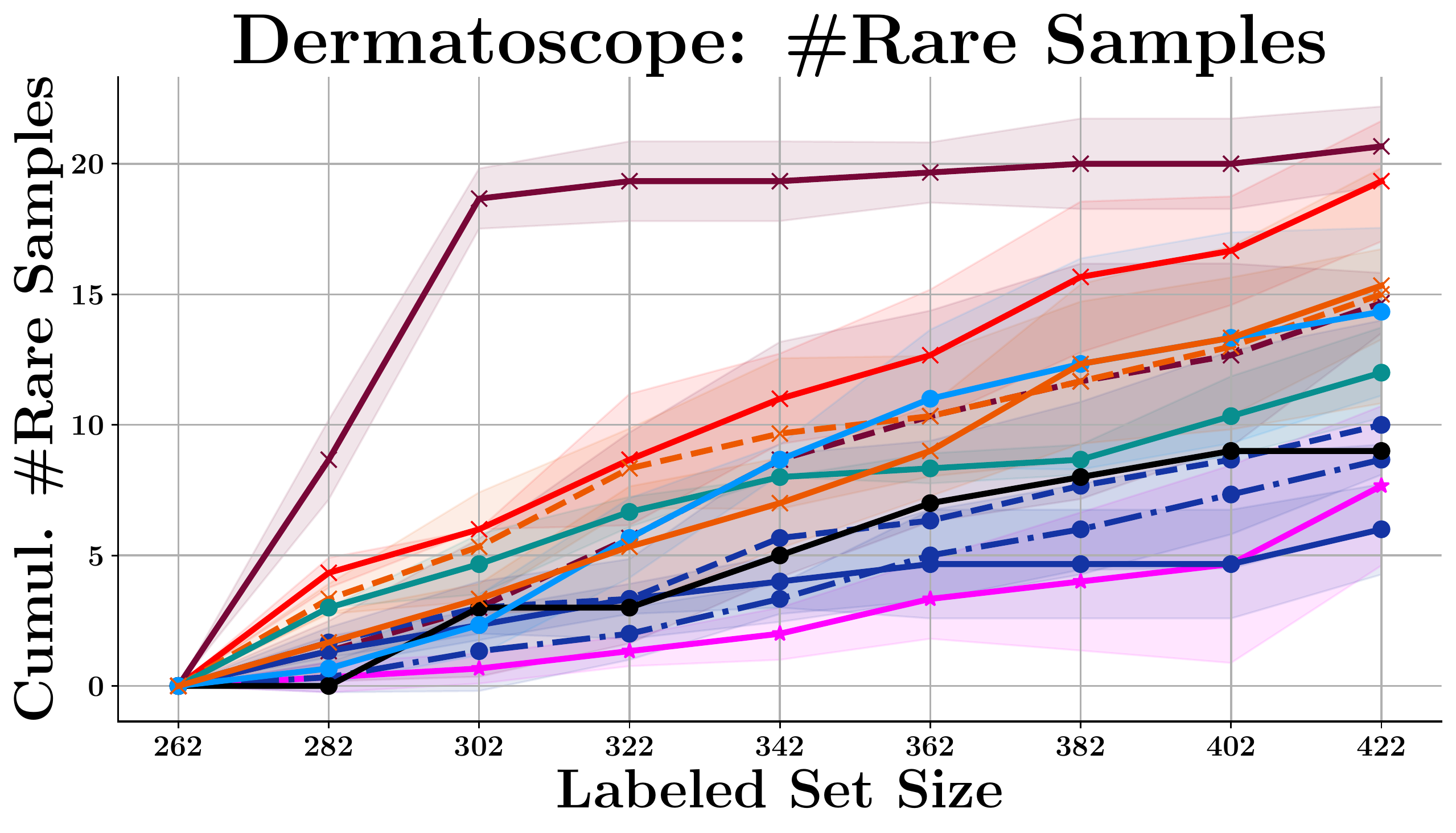}
% \caption{}
\end{subfigure}
\caption{Dermatascope Binary Imbalance Results. We observe that \textsc{Flvmi} and \textsc{Logdetmi} significantly outperform other baselines. Particularly, \textsc{Flvmi} selects significantly more rare class samples than all methods.}
\label{fig:dermamnist_res}
\end{figure*}

We present the results for binary imbalance on the DermaMNIST \cite{codella2019skin} dataset in \figref{fig:dermamnist_res}.

\subsection{Statistical Significance Penalty Matrices} \label{app:pen_matrices}

The penalty matrices computed in this paper follow the strategy used in~\cite{ash2020deep}. In their strategy, a penalty matrix is constructed for each dataset-model pair. Each cell $(i,j)$ of the matrix reflects the fraction of training rounds that AL with selection algorithm $i$ has higher test accuracy than AL with selection algorithm, $j$ with statistical significance. As such, the average difference between the test accuracies of $i$ and $j$ and the standard error of that difference are computed for each training round. A two-tailed $t$-test is then performed for each training round: If $t>t_\alpha$, then $\frac{1}{N_{train}}$ is added to cell $(i,j)$. If $t<-t_\alpha$, then $\frac{1}{N_{train}}$ is added to cell $(j,i)$. Hence, the full penalty matrix gives a holistic understanding of how each selection algorithm compares against the others: A row with mostly high values signals that the associated selection algorithm performs better than the others; however, a column with mostly high values signals that the associated selection algorithm performs worse than the others. As a final note, \cite{ash2020deep} takes an additional step where they consolidate the matrices for each dataset-model pair into one matrix by taking the sum across these matrices, giving a summary of the AL performance for their entire paper that is fairly weighted to each experiment. Below, we present the penalty matrices for each of the settings.

%PATHMNIST and PNEUMONIA MNIST
\begin{figure}[!ht]
\centering
\includegraphics[width =0.9\textwidth]{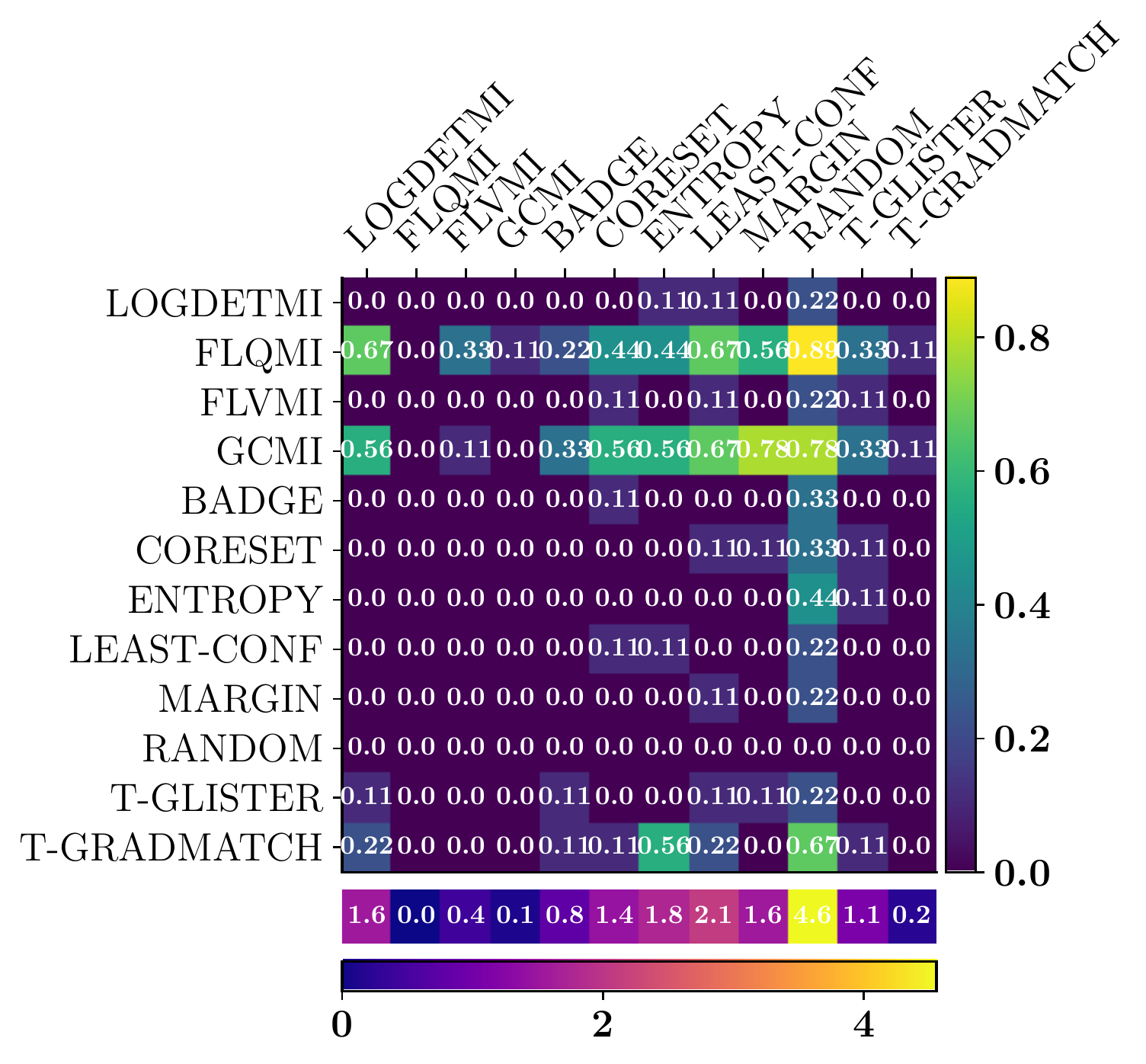}
\includegraphics[width = 0.9\textwidth]{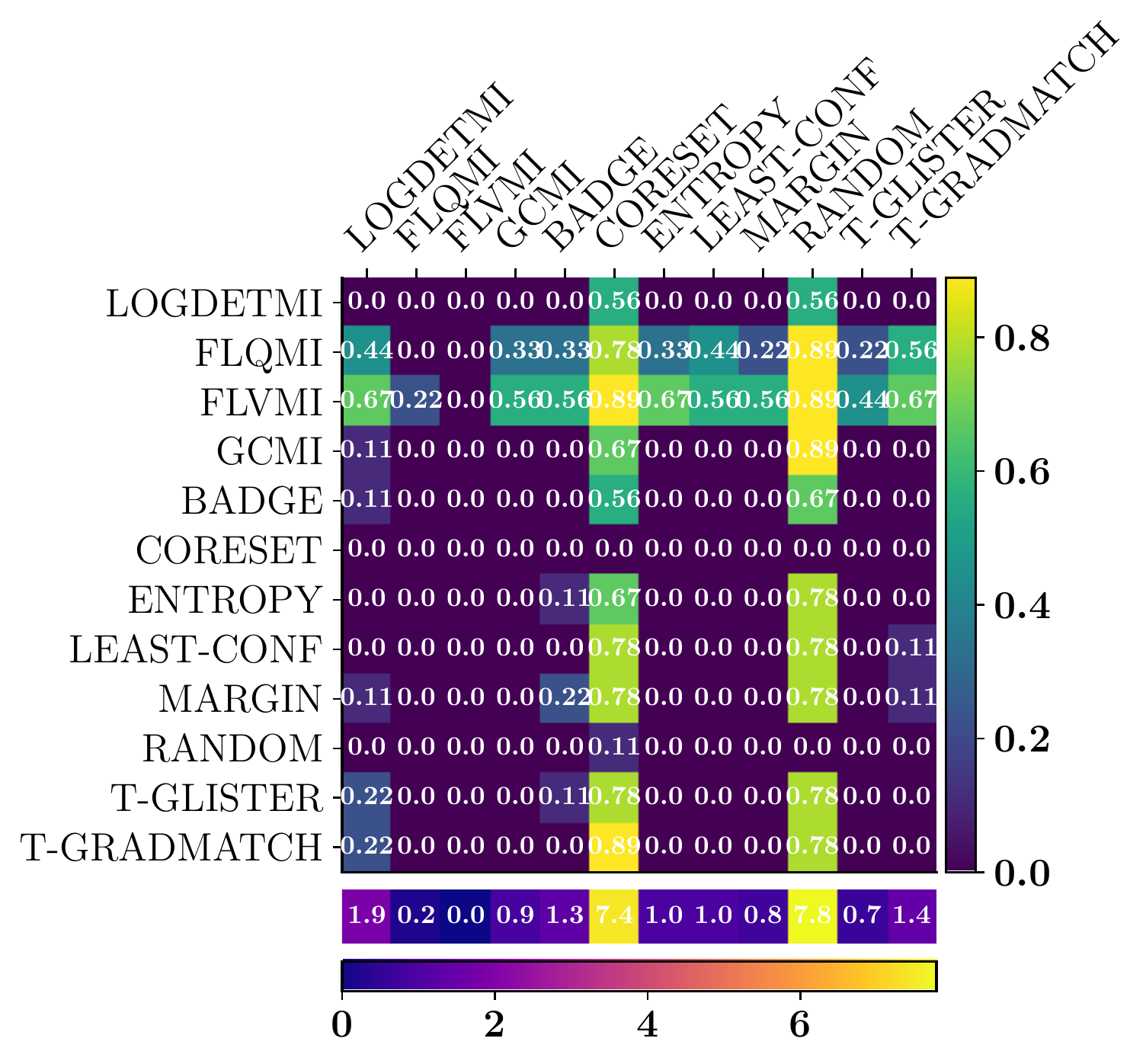}
\caption{Penalty Matrix comparing the average accuracy of binary imbalance for Path-MNIST (\textbf{top}) datasets and Pneumonia-MNIST (\textbf{bottom}) using targeted active learning across multiple runs. We observe that the SMI functions have a much lower column sum compared to other approaches.
}
\label{pen-matrix1}
\end{figure}

\begin{figure}[!ht]
\centering
\includegraphics[width =0.9\textwidth]{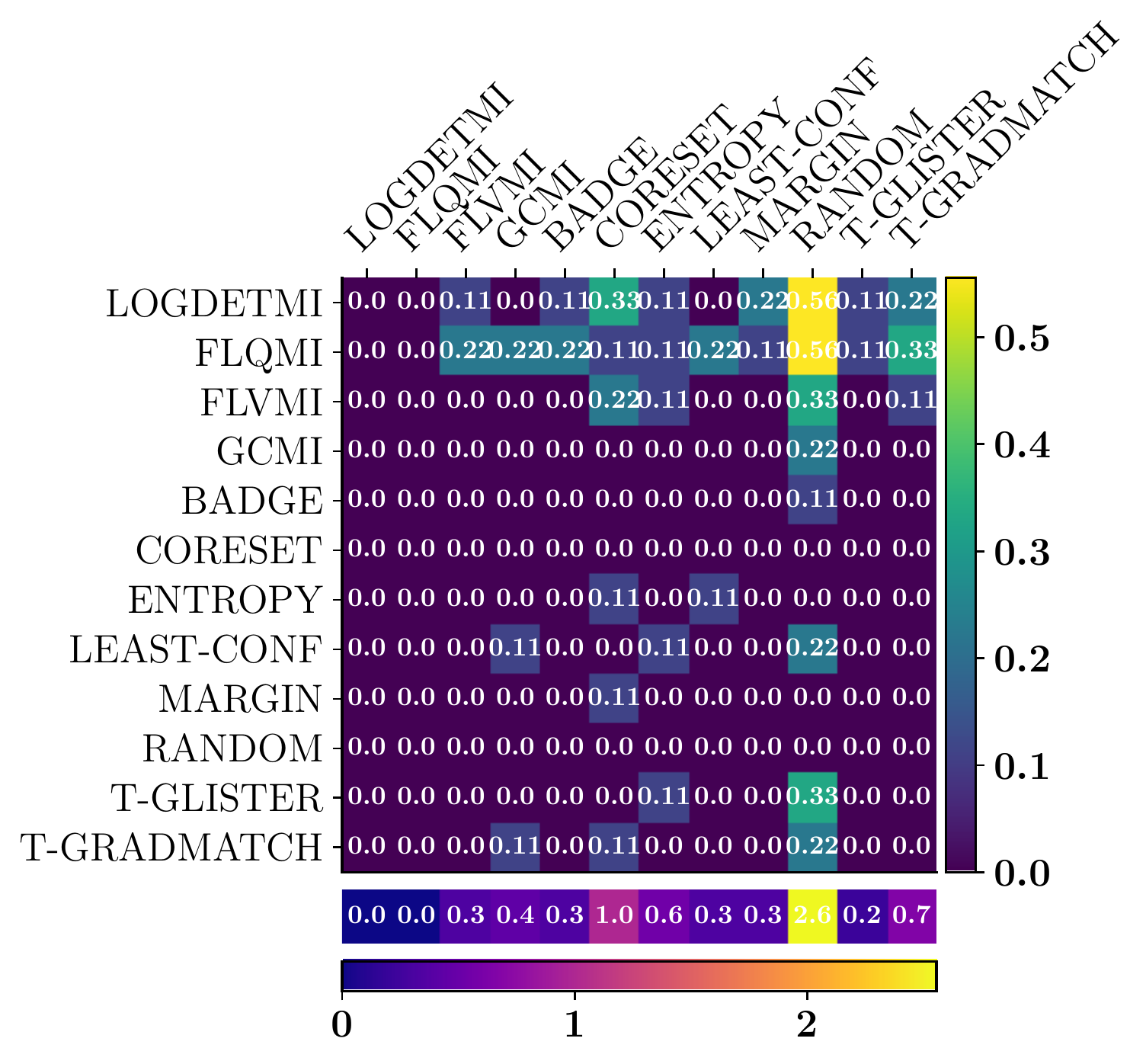}
\includegraphics[width = 0.9\textwidth]{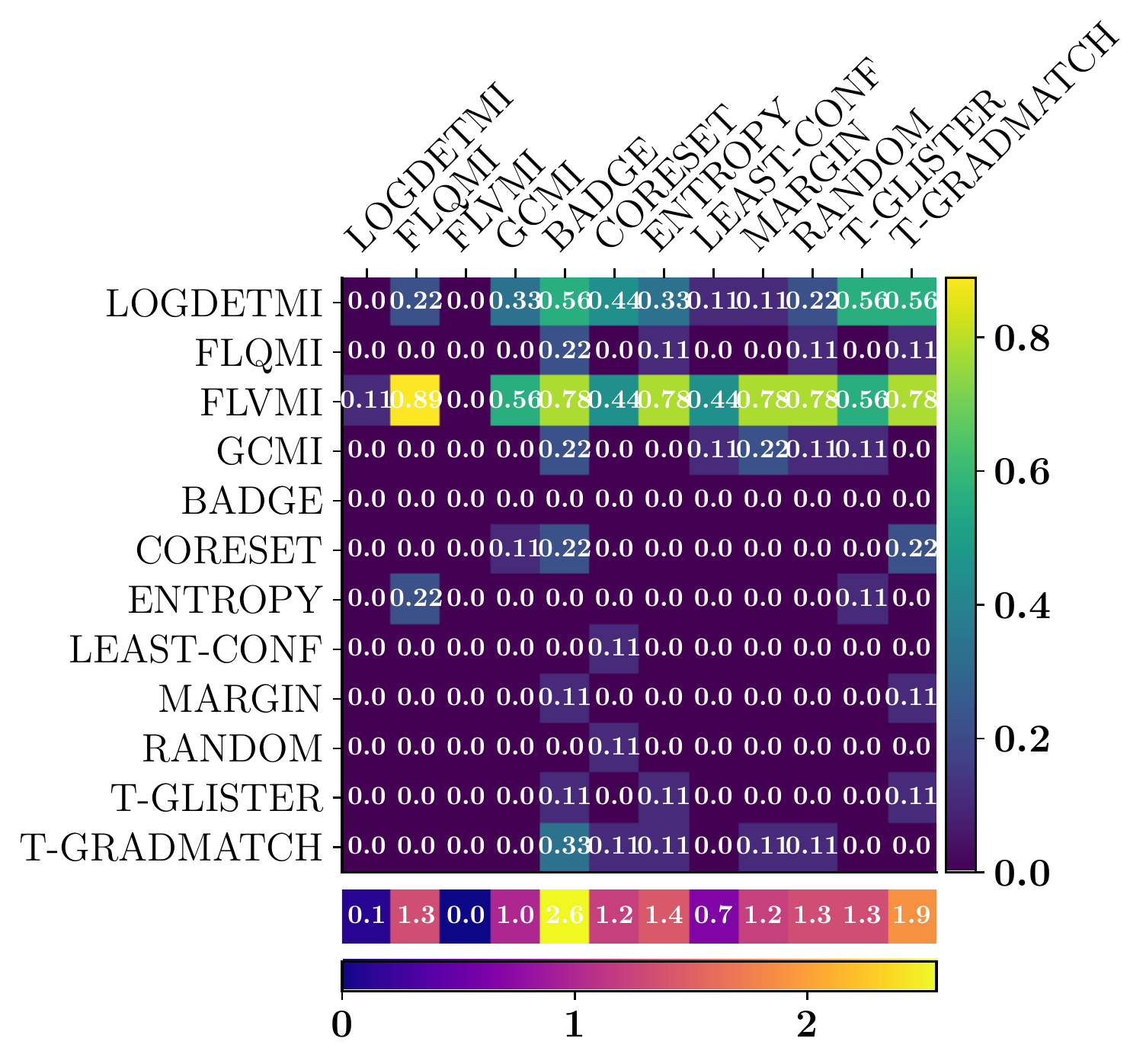}
\caption{Penalty Matrix comparing the average accuracy for rare classes in the binary imbalance for Blood-MNIST (\textbf{top}) and Derma-MNIST (\textbf{bottom}) datasets using targeted active learning across multiple runs. We observe that the SMI functions have a much lower column sum compared to other approaches.
}
\label{pen-matrix2}
\end{figure}

\begin{figure}[!ht]
\centering
\includegraphics[width =0.9\textwidth]{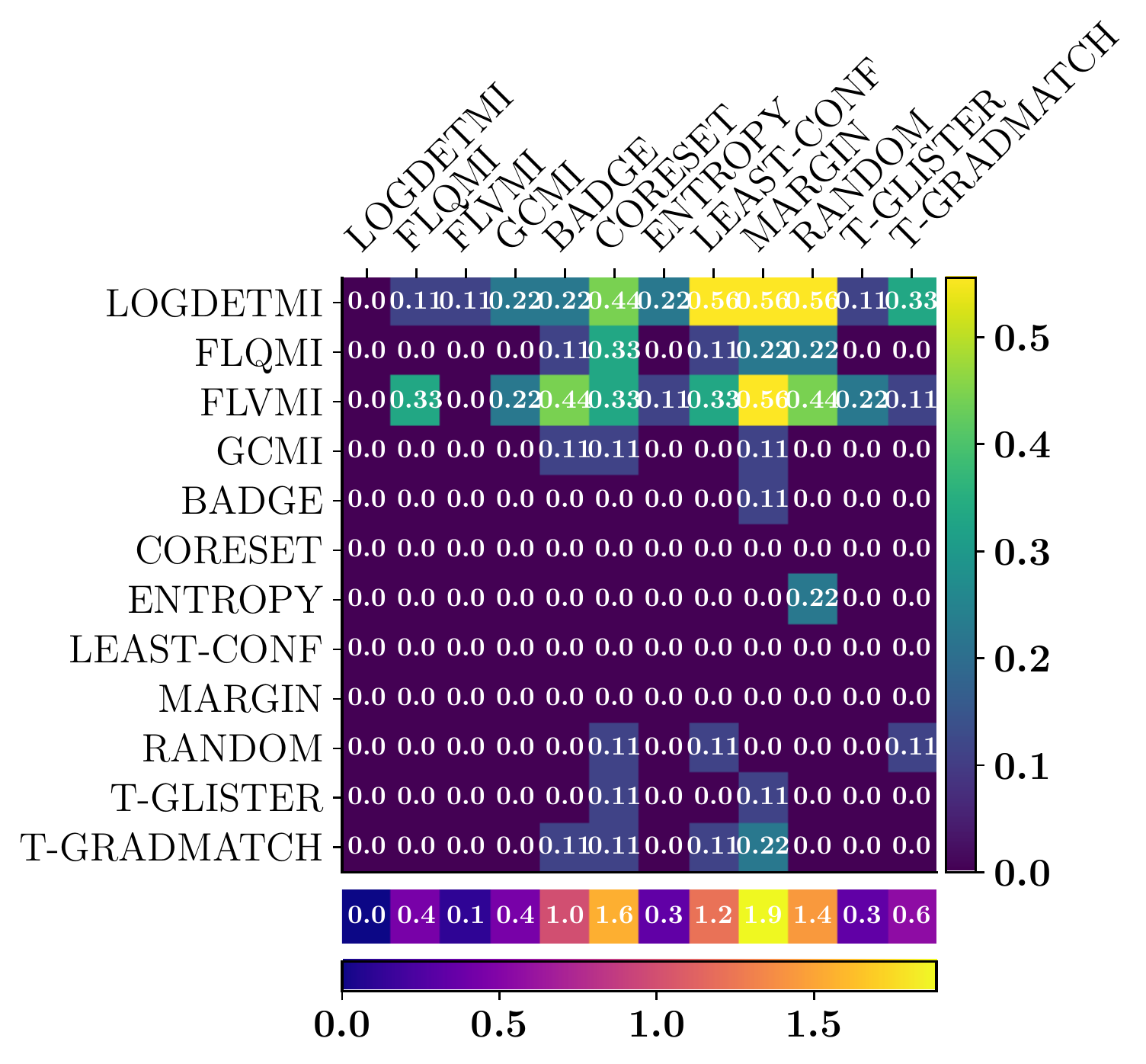}
\includegraphics[width = 0.9\textwidth]{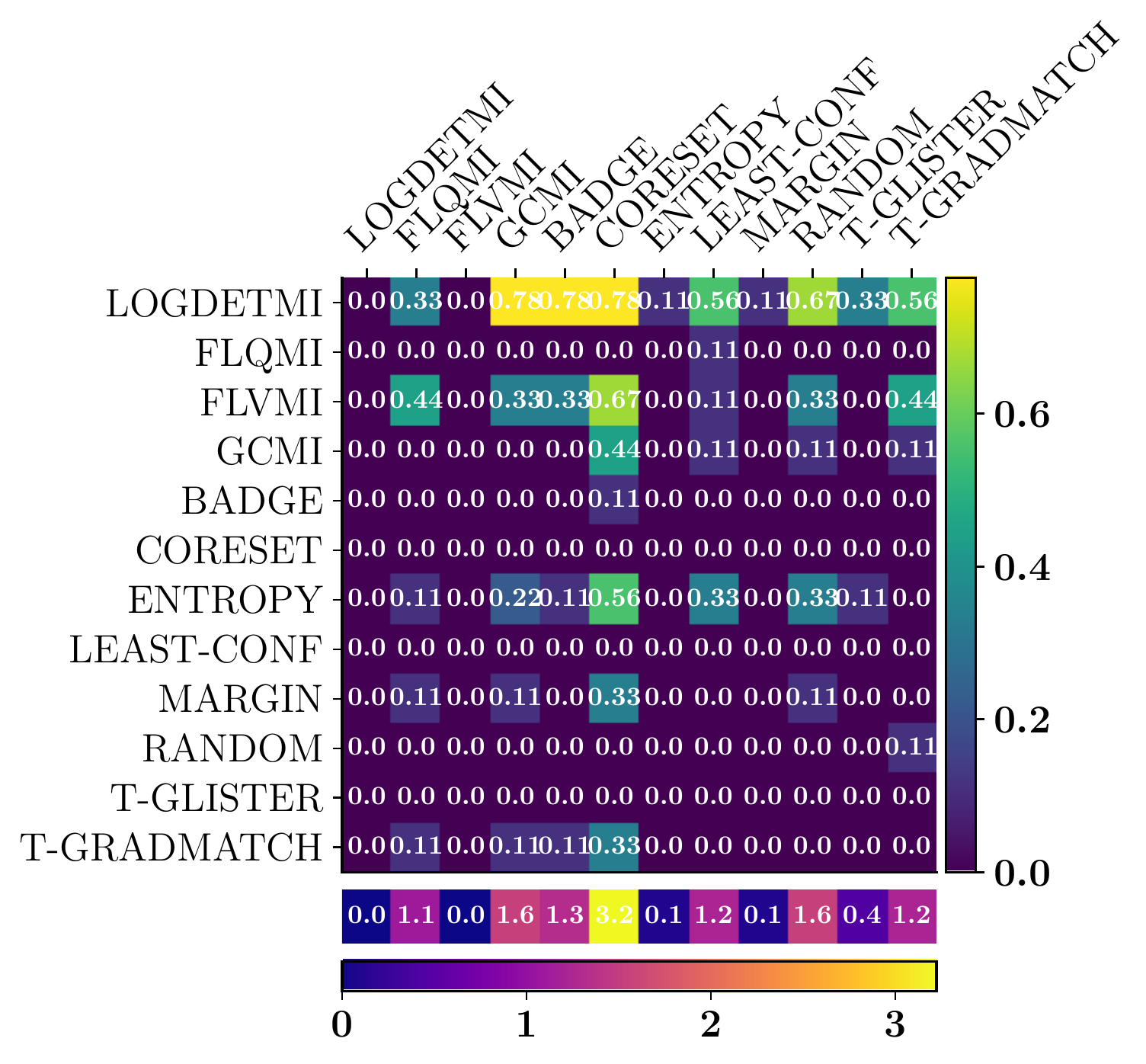}
\caption{Penalty Matrix comparing the average accuracy of rare classes in the long-tail imbalance for ISIC-2018 (\textbf{top}) and APTOS (\textbf{bottom}) datasets using targeted active learning across multiple runs. We observe that the SMI functions have a much lower column sum compared to other approaches.
}
\label{pen-matrix3}
\end{figure}

\end{document}